\documentclass[conference]{IEEEtran}
\IEEEoverridecommandlockouts

\usepackage{cite}
\usepackage{amsmath,amssymb,amsfonts}
\usepackage{algorithmic}
\usepackage{graphicx}
\usepackage{textcomp}
\usepackage{xcolor}
\def\BibTeX{{\rm B\kern-.05em{\sc i\kern-.025em b}\kern-.08em
    T\kern-.1667em\lower.7ex\hbox{E}\kern-.125emX}}
\usepackage{soul}
\usepackage{pifont}
\usepackage{physics}
\usepackage{tikz,pgfplots}
\usepackage{amsmath,amsfonts}
\usepackage{amsthm}
\usepackage{mathtools}
\usepackage{graphicx}
\usepackage{booktabs}
\usepackage{algorithm}
\usepackage{algorithmic}
\usepackage{makecell}
\usepackage{multirow}
\usepackage[hidelinks]{hyperref}
\usepackage{url}
\usepackage{colortbl}
\usepackage{array}
\usepackage[font=small]{caption}
\usepackage[font=footnotesize]{subcaption}
\usepackage[stable]{footmisc}
\theoremstyle{definition}

\usepackage{mathrsfs}
\usepackage{enumitem}
\usepackage{romannum}
\usepackage{multicol}
\pgfplotsset{compat=1.18}

\usepackage{newfloat}
\DeclareFloatingEnvironment[name={Revision Figure}]{rebuttalfigure}
\DeclareFloatingEnvironment[name={Revision Table}]{rebuttaltable}

\renewcommand*{\arraystretch}{1}

\newcolumntype{P}[1]{>{\centering\arraybackslash}p{#1}}

\makeatletter
\def\footnoterule{\kern-3\p@
  \hrule \@width 2in \kern 2.6\p@} % the \hrule is .4pt high
\makeatother

\usepackage[english]{babel}
\usepackage{xspace}

\newcommand{\framework}{\texttt{EDAD}\xspace}
\newcommand{\cell}{\cellcolor{blue!15}}

\begin{document}

\title{An Encode-then-Decompose Approach to Unsupervised Time Series Anomaly Detection on Contaminated Training Data--Extended Version}

\author{
    \IEEEauthorblockN{
        Buang Zhang$^{1}$, 
        Tung Kieu$^{2}$, 
        Xiangfei Qiu$^{1}$,
        Chenjuan Guo$^{1}$,
        Jilin Hu$^{1}$ \\
        Aoying Zhou$^{1}$,
        Christian S. Jensen$^{2}$,
        Bin Yang$^{1}$ 
    } 
    \IEEEauthorblockA{
        $^1$\textit{School of Data Science \& Engineering, East China Normal University, Shanghai, China} \\ $^2$\textit{Department of Computer Science, Aalborg University, Aalborg, Denmark} \\
        $^1$\{buazhang, xfqiu\}@stu.ecnu.edu.cn, 
        $^1$\{cjguo, jlhu, ayzhou, byang\}@dase.ecnu.edu.cn,
        $^2$\{tungkvt,csj\}@cs.aau.dk
    }
}   

\maketitle
\begin{abstract}
Time series anomaly detection is important in modern large-scale systems and is applied in a variety of domains to analyze and monitor the operation of diverse systems. 
Unsupervised approaches have received widespread interest, as they do not require anomaly labels during training, thus avoiding potentially high costs and having wider applications. 
Among these, autoencoders have received extensive attention. 
They use reconstruction errors from compressed representations to define anomaly scores. 
However, representations learned by autoencoders are sensitive to anomalies in training time series, causing reduced accuracy. 
We propose a novel encode-then-decompose paradigm, where we decompose the encoded representation into stable and auxiliary representations, thereby enhancing the robustness when training with contaminated time series. 
In addition, we propose a novel mutual information based metric to replace the reconstruction errors for identifying anomalies. 
Our proposal demonstrates competitive or state-of-the-art performance on eight commonly used multi- and univariate time series benchmarks and exhibits robustness to time series with different contamination ratios.
\end{abstract}

\thispagestyle{plain}
\pagenumbering{arabic}
\pagestyle{plain}

\section{Introduction}
\label{sec:introduction3}

Time-ordered data, known as time series, from a variety of embedded sensors has become the foundation for the continuous monitoring and management of large-scale systems across a variety of domains such as healthcare~\cite{DBLP:conf/kdd/WangLYY023}, finance~\cite{DBLP:conf/aistats/BachelardCFT23}, logistics~\cite{DBLP:conf/cikm/YiYWYL23}, manufacturing~\cite{DBLP:conf/cikm/Wang0PB16}, and natural sciences~\cite{DBLP:conf/sigmod/LernerSWZZ04}.
Time series anomaly detection, an important branch of time series analysis, constitutes fundamental functionality in data analytics, data management, and data mining. 
Time series anomaly detection is receiving increasing attention in academia and industry, with numerous applications that include system maintenance~\cite{DBLP:conf/kdd/RamakrishnanSLS19}, network intrusion monitoring~\cite{DBLP:conf/ccs/SekarGFSTYZ02}, and credit card fraud detection~\cite{DBLP:journals/pvldb/XiaoW00O23}.
The lack of labeled data and the diversity of anomalies combine to make the problem of identifying anomalies challenging and to limit the applicability of methods that require supervision. 
This has spurred research on unsupervised methods, leading to promising results.

\begin{figure}[t]
    \centering
    \includegraphics[width=1.0\linewidth]{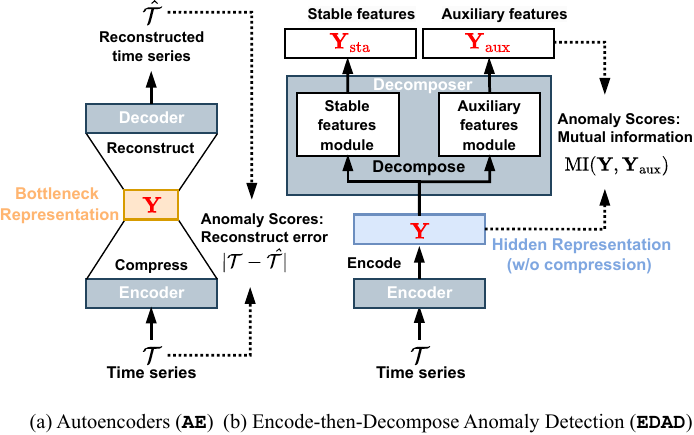}
    \caption{Autoencoders (\texttt{AE}) vs. Encode-then-Decompose Anomaly Detection (\texttt{EDAD}).}
    \label{fig:comparison_ae_edad}
    % \vspace{-1.5em}
\end{figure}

Recent neural network based methods for time series anomaly 
detection achieve strong performance on challenging datasets~\cite{DBLP:conf/ijcai/KieuYGJ19}. These methods are able to learn long-term, nonlinear temporal relationships in the data, outperforming existing “shallow” methods based on similarity search~\cite{DBLP:conf/icde/BoniolLRP20a,DBLP:journals/pvldb/BoniolPKPTEF22,DBLP:conf/edbt/Senin0WOGBCF15} and density-based clustering~\cite{DBLP:conf/sigmod/BreunigKNS00}. Among the neural network based methods, a commonly used paradigm adopts an encoder-decoder mechanism, that first compresses time series into a compact, hidden representation, and then reconstructs the time series from the hidden representation, as illustrated in Figure~\ref{fig:comparison_ae_edad}(a). This paradigm employs a so-called autoencoder (\texttt{AE})~\cite{DBLP:journals/corr/abs-2201-03898}, which imposes an information bottleneck~\cite{DBLP:journals/corr/abs-2310-03311} that encourages the compact latent representation to capture only the most representative patterns of the input time series, while disregarding fluctuations in the time series.
Although autoencoders achieve impressive accuracy, they face the following two limitations.  

\noindent\textbf{Compress-then-Reconstruct paradigm: } 
\texttt{AE}s employ a Compress-then-Reconstruct paradigm, as shown in Figure~\ref{fig:comparison_ae_edad}(a).
The training time series $\mathcal{T}$ are often required to be fully clean, i.e., without anomalies, such that the bottleneck representation captures the most essential, normal patterns. 
When the training time series includes anomalies, they may pollute the bottleneck representation such that it also captures anomalous patterns, thus adversely affecting anomaly detection, i.e., causing some anomalies to have small reconstruction errors. 
A more robust paradigm that is able to better deal with contaminated training data is desirable.

% \vspace{0.5em}
\noindent
\textbf{Symmetric design of loss functions and anomaly scores: } 
The Compress-then-Reconstruct paradigm often uses a symmetric design of the training loss functions and anomaly scores, i.e., both rely on reconstruction errors. This works well if the training data is clean. 
However, this symmetric design is problematic when training with contaminated time series. 
Specifically, during training, we still aim to minimize the reconstruction errors between the input time series $\mathcal{T}$ and the reconstructed time series $\hat{\mathcal{T}}$. 
If $\mathcal{T}$ already includes anomalies, minimizing the training loss drives the autoencoder to learn a bottleneck representation that also captures anomalous patterns caused by anomalies.  
Thus, in the testing phase, the reconstruction errors for some anomalies may be small and thus difficult to detect. 
To conclude, this symmetric design causes a \textcolor{black}{problem}---training with contaminated data reduces the detection accuracy.   
This calls for means to avoid this problem. 

To address the two limitations, we propose an Encode-then-Decompose Anomaly Detection (\texttt{EDAD}) framework. \texttt{EDAD} employs a novel ``Encode-then-Decompose'' paradigm with an asymmetric design of loss functions and anomaly scores, where effective mutual information based metrics are proposed to enhance the robustness w.r.t. contaminated training data.

\begin{table}[t]
    % \small
    \footnotesize
    % \fontsize{7.5pt}{7.5pt}\selectfont
    \tabcolsep2.5pt
    \centering
    \caption{Comparison of Autoencoder vs. Encode-then-Decompose Anomaly Detection, where MI denotes mutual information.}
    \label{tab:comparison_intro}
    % 增加全局行高
    \renewcommand{\arraystretch}{1.5}
    \begin{tabular}{l|l|l}
        \toprule
        \textbf{} & \textbf{AutoEncoder } & \textbf{EDAD} \\
        \midrule
        \textbf{Paradigm} & Compress-then-Reconstruct & Encode-then-Decompose \\ 
        \textbf{Outlier Scores} & Reconstruction Errors & MI($\mathbf{Y}$, $\mathbf{Y}_{\text{aux}}$) \\ 
            \textbf{Training Loss} &  Reconstruction Errors & \parbox{3cm}{MI($\mathbf{Y}$, $\mathbf{Y}_{\text{sta}}$) + Closeness} \\
        \textbf{Training Data} & Clean Time Series& Contaminated Time Series \\
        \bottomrule
    \end{tabular}
    % \vspace{-1.5em}
\end{table}
% \vspace{-0.5em}

\noindent\textbf{Encode-then-Decompose paradigm: } 
We propose an  \textit{Encode-then-Decompose} paradigm that aims to improve robustness to training with contaminated time series data.
Instead of using a single bottleneck representation to capture the information of input time series, we decompose a single representation into two---one representing stable patterns and the other representing auxiliary patterns. 
% The stable patterns capture the main features in the time series, while the auxiliary patterns represent anomalies or noise components. 
This design aims to separate abnormal patterns in contaminated time series from normal patterns, to achieve better robustness than the Compress-then-Reconstruct paradigm.

The proposed decomposition occurs in the latent representation space, which we call a ``deep'' decomposition, whereas existing time series decompositions often work on the time series themselves, which we refer to as ``shallow'' decompositions~\cite{west1997time,cleveland1976decomposition,theodosiou2011forecasting}. 
Specifically, deep decomposition separates the encoded latent representation into two components: \textit{stable} features and \textit{auxiliary} features. The stable features capture shared, invariant patterns across the time series, while the auxiliary features reflect local variations and noise. 
Importantly, the latent space--constructed through attention modules with linear embedding layers--preserves the original temporal dependencies~\cite{DBLP:conf/aaai/ZengCZ023}. The proposed deep decomposition is achieved by a novel design of shuffle strategies along the time dimension, i.e., randomly changing the time order of the elements in learned representations. Consequently, shuffling the order of data points in this latent space effectively corresponds to shuffling their order in the original data, albeit indirectly.
More specifically, the features that are insensitive to shuffling are stable features, whereas the features that are sensitive to shuffling are auxiliary features. 
This implies that stable features exhibit consistent patterns over time and are not prone to unpredictable fluctuations.
In contrast, auxiliary features are sensitive to temporal order, making them effective for capturing localized, short-term patterns, and noise in the time series.
This design is fully \textit{unsupervised} and \textit{parameter-free}, thus enabling unsupervised anomaly detection when training with unlabeled, contaminated time series data. 

% \vspace{0.5em}
\noindent
\textbf{Asymmetric design of loss functions and anomaly scores: } 
The proposed Encode-then-Decompose paradigm's decomposition of time series representations into stable features and auxiliary features facilitates an asymmetric design. Instead of using reconstruction errors, we use mutual information as a novel and important metric when designing the training loss and computing anomaly scores.
During training, we consider two aspects to guide the framework's learning. First, the auxiliary representation $\mathbf{Y}_{\text{aux}}$, which represents point-wise features, is sensitive to shuffling, and the stable representation $\mathbf{Y}_{\text{sta}}$, which represents long-term features such as trend and seasonalities, is insensitive to the shuffling. 
Second, the stable representation, $\mathbf{Y}_{\text{sta}}$, and the original hidden representation before decomposition, $\mathbf{Y}$, have large mutual information. 
This is because the stable representation $\mathbf{Y}_{\text{sta}}$ captures the majority of the normal patterns in $\mathbf{Y}$ according to our definition of \textit{stable}.
During testing, we use the point-wise mutual information between $\mathbf{Y}_{\text{aux}}$ and $\mathbf{Y}$ to obtain anomaly scores because $\mathbf{Y}_{\text{aux}}$ captures unexpected variations in time series.
% specific variations at a particular point. 
If $\mathbf{Y}$ and $\mathbf{Y}_{\text{aux}}$ have low mutual information, a time series point is likely to be an anomaly. In summary, the Encode-then-Decompose paradigm facilitates separation between training loss and anomaly scores, thus enabling an asymmetric design.

Table~\ref{tab:comparison_intro} summarizes key differences between the existing vs. the proposed paradigm. 
To the best of our knowledge, this is the first study to propose a deep decomposition paradigm for unsupervised time series anomaly detection using mutual information. 
In summary, the contributions of the paper are as follows. (i) We propose a novel Encode-then-Decompose paradigm to distinguish between long-term patterns (\textit{stable} features) and short-term patterns (\textit{auxiliary} features), thus mitigating the negative effects of training on contaminated data. (ii) We propose a latent space point-wise mutual information criterion for anomaly detection and form an asymmetric pipeline with a decomposition framework to improve robustness. We also introduce a novel loss function to train the framework using mutual information. (iii) We report on extensive experiments on eight benchmark datasets using multiple metrics to assess the effectiveness of the proposal and offer detailed insight into its performance characteristics.
% \begin{itemize}

%     \item  We propose a novel Encode-then-Decompose paradigm to distinguish between long-term patterns (\textit{stable} features) and short-term patterns (\textit{auxiliary} features), thus mitigating the negative effects of training on contaminated data.     
%     \item We propose a latent space point-wise mutual information criterion for anomaly detection and form an asymmetric pipeline with a decomposition framework to improve robustness. We also introduce a novel loss function to train the framework using mutual information. 
%     \item We report on extensive experiments on eight benchmark datasets using multiple metrics to assess the effectiveness of the proposal and offer detailed insight into its performance characteristics.
% \end{itemize}

The rest of the paper is organized as follows. 
Section~\ref{sec:pre} covers preliminaries. %
Section~\ref{sec:method} details the proposal.
Section~\ref{sec:exp} reports on the experimental study, Section~\ref{sec:related} covers related work, and Section~\ref{sec:con} concludes.

\section{Preliminaries}
\label{sec:pre}
\subsection{Time Series} 
A time series $\mathcal{T} = \langle \mathbf{s}_{1}, \mathbf{s}_{2}, \dots, \mathbf{s}_{N} \rangle$ is a sequence of $N$ time-ordered observations, where each observation $\mathbf{s}_i\in\mathbb{R}^{D}$ is collected at a specific time step. If $D=1$, $\mathcal{T}$ is \textit{univariate}. If $D>1$, $\mathcal{T}$ is \textit{multivariate} (or \textit{multidimensional}).

\subsection{Time Series Anomaly Detection}
\label{subsec:time-series-anomaly-detection}
Given a time series $\mathcal{T} = \langle \mathbf{s}_{1}, \mathbf{s}_{2}, \dots, \mathbf{s}_{N} \rangle$, we aim at computing an anomaly score $\mathcal{AS}(\mathbf{s}_{i})$ for each observation $\mathbf{s}_{i}$ such that the higher $\mathcal{AS}(\mathbf{s}_{i})$ is, the more likely it is that  $\mathbf{s}_{i}$ is an anomaly. We focus on the unsupervised anomaly detection problem, as no labels (neither for anomalies nor for normal data) are used during training. This follows the definition of ``unsupervised" commonly adopted in prior studies~\cite{DBLP:conf/ijcai/KieuYGJ19,DBLP:conf/kdd/SuZNLSP19,DBLP:conf/www/XuCZLBLLZPFCWQ18}. In contrast, semi-supervised anomaly detection assumes access to a small number of labeled normal and/or anomalous instances, which is not the case in our work. Further, we make no assumptions about whether anomalies are point or collective anomalies. If the anomaly scores of continuous observations are high, these observations can be considered as a collective anomaly. 
As discussed in Section~\ref{sec:introduction3}, in the Compress-then-Reconstruct paradigm, reconstruction errors are used as anomaly scores; in the proposed Encode-then-Decompose paradigm, latent space point-wise mutual information between an encoded representation and auxiliary features is used for defining anomaly scores, which we will detail in Section~\ref{sec:method}. 

\subsection{Mutual Information Estimation for High-Dimensional Data} 

Mutual information (MI) measures the statistical dependency between random variables. 
Formally, given random variables $X$ and $Y$, the MI between $X$ and $Y$, denoted as $I(X,Y)$, is defined as follows.

% \vspace{-0.5em}
\begin{equation}
    % \small
    I(X,Y) = \sum_{x\in X} \sum_{y\in Y}
    \mathbb{P}(x,y)\log\left(\frac{\mathbb{P}(x,y)}{\mathbb{P}(x)\mathbb{P}(y)}\right)
    \label{equation:mutual_information}
\end{equation}
% \vspace{-0.5em}

%
\noindent Here, $\mathbb{P}(x,y)$ indicate the joint distribution, and $\mathbb{P}(x)$ and $\mathbb{P}(y)$ are the marginal distributions of $X$ and $Y$ obtained through a marginalization process. Note that in the context of time series, both $X$ and $Y$ are continuous variables.

Conceptually, MI quantifies the amount of shared information between a pair of random variables, which measures the uncertainty in one variable if the knowledge of the other variable is provided, and vice versa. 
In other words, the higher the MI value is, the more information the two random variables share---knowing one random variable thus reduces the uncertainty of the other random variable to a large extent. 
In contrast, if random variables $X$ and $Y$ are independent, they do not share any information, and knowing one random variable does not reduce the uncertainty of the other random variable, thus making their MI equal to 0. 

In this paper, we need to compute the mutual information between timestamps of time series in a latent space. Generally, the representation of timestamps in the latent space can be considered as a high-dimensional vector. Classical mutual information estimation methods are intractable for such vectors~\cite{Kraskov_Stögbauer_Grassberger_2004}. The estimation of mutual information on large-scale data or high-dimensional variables remains challenging.

With the recent advances in mutual information estimation,  accurate estimators of mutual information between high-dimensional variables are available. By introducing variational bounds and inequalities, the problem of directly estimating density ratios has been transformed into estimating an optimization problem.

Specifically, we can use the following unnormalized version of the Barber and Agakov approximation $I_{\mathrm{UBA}}(X, Y)$ to approximate the mutual information $I(X, Y)$ between random variables $X$ and $Y$~\cite{DBLP:conf/icml/PooleOOAT19}. 

% \vspace{-0.5em}
\begin{equation}
    % \small
    \label{eq:iuba}
    \begin{aligned}
        I_{\mathrm{UBA}}(X, Y)& \triangleq\mathbb{E}_{p(x,y)}[\log q(x|y)]+h(X)  \\
        &=\mathbb{E}_{p(x,y)}[\log p(x)-\log Z(y)+f(x,y)]+h(X) \\
        &=\mathbb{E}_{p(x,y)}[f(x,y)]-\mathbb{E}_{p(y)}[\log Z(y)]
    \end{aligned}
\end{equation}
% \vspace{-1.0em}

\noindent Here, $h(X)=-\mathbb{E}_{p(x)}\left[\log(p(x))\right]$ is  the differential entropy of $X$, $Z(y)=E_{p(x)}\left[e^{f(x,y)}\right]$, and  $q(x|y)$ denotes the conditional probability of $X$ given $Y$, which is defined as follows.

% \vspace{-0.5em}
\begin{equation}
    % \small
    \label{eq:energy}
    q(x|y) = \frac{p(x)}{Z(y)}e^{f(x,y)}
\end{equation}
% \vspace{-1.0em}

\noindent Here, $q(x|y)$ is considered as an energy function in  system~\cite{Brereton_1981}, $e^{f(x,y)}$ is a tilting function,  $f(x,y):\mathcal{X}\times\mathcal{Y}\to\mathbb{R}$ is a critic function aiming to distinguish whether the $x$ and $y$ come from the same joint distribution, and $Z(y)$ is the associated partition function.

If we use different techniques to deal with the factor in  Equation~\ref{eq:iuba}, we get a variety of different variational mutual information estimators, including \texttt{MINE}~\cite{DBLP:conf/icml/BelghaziBROBHC18}, \texttt{NWJ}~\cite{DBLP:conf/icml/BelghaziBROBHC18}, \texttt{InfoNCE}~\cite{DBLP:journals/corr/abs-1807-03748}, and \texttt{JSD}~\cite{DBLP:conf/iclr/HjelmFLGBTB19}.

\section{Methodology}
\label{sec:method}
We first present an overview of the Encode-then-Decompose Anomaly Detection (\texttt{EDAD}) framework that efficiently decomposes a learned hidden time series representation into stable and auxiliary representations.
Next, we present the objective function, which is based on representation closeness and mutual information. 
This function aims to enable robust training, to contend settings with contaminated training data.

\subsection{Framework Overview}
\label{subsec:framework-overview}

\begin{figure}[h]
	\centering
	\includegraphics[width=0.9\linewidth]{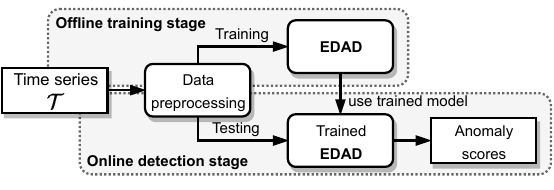}
	\caption{Framework pipeline. The data preprocessing component is shared by the offline training and online detection stages.}
	\label{fig:framework}
	% \vspace{-0.5em}
\end{figure}

An overview of the framework is shown in Figure~\ref{fig:framework}. 
The proposed framework consists of two stages, covering offline training and online detection. In the offline training stage, the model training is performed on time series datasets that may already include anomalies. In the online detection stage, the trained model is used for detecting anomalies. 

The data preprocessing component is shared by the offline training and online detection. 
This component adopts an established technique~\cite{DBLP:conf/kdd/SuZNLSP19,DBLP:conf/icde/KieuYGCZSJ22} and applies the dimension independence strategy, which is the state-of-the-art method for time series~\cite{DBLP:conf/aaai/ZengCZ023}. This strategy assumes that the dimensions of a time series do not share information. Thus, it disregards correlations between dimensions. When applying the dimension independence strategy, the model is forced to capture long-term temporal dependencies within each channel and preventing it from trivially inferring a variable's value based solely on other channels. Prior studies~\cite{DBLP:conf/iclr/NieNSK23,DBLP:conf/iclr/LiuHZWWML24} have also reported that the independence-channel setting usually outperforms cross-channel modeling. In other words, the dimension independence strategy can be considered as a consolidation and temporal augmentation method. We apply the dimension independence strategy as follows. 
A multivariate time series $\mathcal{T} \in \mathbb{R}^{N \times D}$ is treated as $D$ univariate time series $\mathcal{T}_{j} \in \mathbb{R}^{N \times 1}, j = 1, 2, \ldots, D$.
The univariate time series are each standardized and partitioned into overlapping subsequences by using a sliding window of length $B$. 
Then, the resulting sequences of length $B$, are fed into the model for training. 
Here, we propose \texttt{EDAD}, which is explained in the following parts.
After training, the learned models are then employed for online anomaly detection. 
Specifically, each sequence is preprocessed by the data preprocessing component and then fed into the trained \texttt{EDAD} model that outputs an anomaly score for each observation in the series.

% \subsection{Motivation}

\subsection{Network Architecture}
\begin{figure*}[t]
    \centering
    \includegraphics[width=0.95\linewidth]{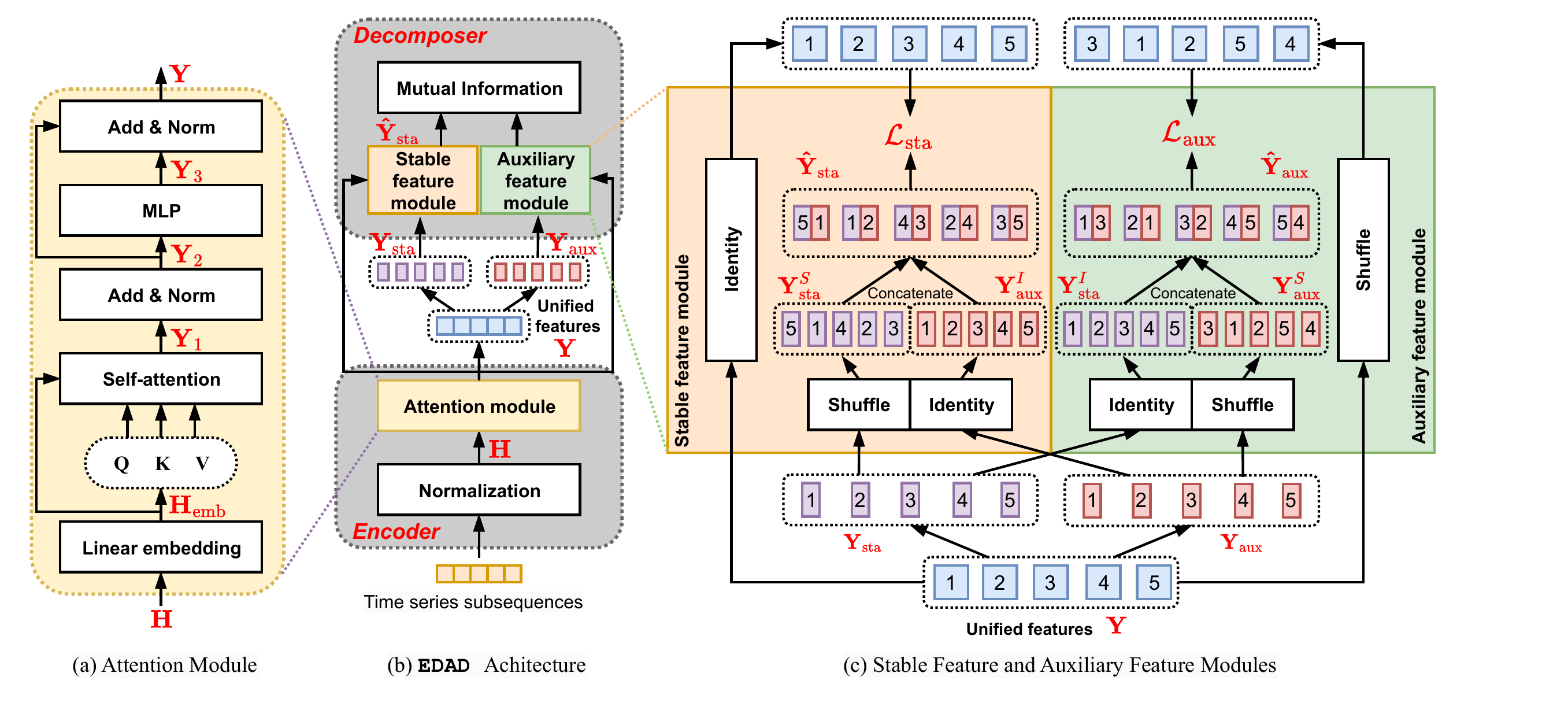}
    \caption{\texttt{EDAD} overview.} 
    \label{fig:architecture}
    % \vspace{-1.5em}
\end{figure*}

We propose a novel 
\textcolor{black}{\textit{encoder-decomposer} based} architecture as the backbone of the \texttt{EDAD} framework, as illustrated in Figure~\ref{fig:architecture}.

% \begin{figure}[t]
%     \centering
%     \includegraphics[width=0.85\linewidth]{figures/architecture_v2.pdf}
%     \caption{\texttt{EDAD} Architecture.} 
%     \label{fig:architecture}
% \end{figure}

% \begin{figure}[h]
% 	\centering
% 	\includegraphics[width=0.9\linewidth]{mimodule.png}

%         \caption{MI Module} 
% 	\label{fig:framework}
% 	\vspace{-1.0em}
% \end{figure}

The framework comprises two components---an \textit{encoder} and a \textit{decomposer}.
The encoder encompasses an  \textit{attention module}. 
The decomposer encompasses two modules--- \textit{stable feature module} and  \textit{auxiliary feature module}. 
The preprocessed time series are input into a normalizing layer to perform instance normalization~\cite{DBLP:journals/corr/UlyanovVL16}, defined as follows.

% \vspace{-0.5em}
\begin{equation}
    % \small
    \mathbf{H}_{t:t+B} = \frac{\mathbf{s}_{t:t+B} - \mathbb{E}[\mathbf{s}_{t:t+B}]}{\sqrt{\mathrm{Var}[\mathbf{s}_{t:t+B}] + \epsilon}} \cdot \gamma_{1} + \beta_{1}
    \label{equation:norm}
\end{equation}
% \vspace{-0.5em}

\noindent Here, the $\gamma_{1}$ and $\beta_{1}$ are learnable parameter vectors, and $\mathbb{E}[\mathbf{s}_{t:t+B}]$ and $\mathrm{Var}[\mathbf{s}_{t:t+B}]$ are the expectation and variance of a time series subsequences, respectively. 
\textcolor{black}{The output of Equation~\ref{equation:norm} for $\mathbf{s}_{t:t+B}$ is $\mathbf{H}_{t:t+B}$. 
However, for simplicity, we omit ${t:t+B}$ in the following.}

\subsubsection{Attention Module}
The reason for using attention mechanisms is twofold. First, attention mechanisms offer high parallelism and the ability to capture long-range dependencies. Second, in contrast to \texttt{AE} with compression mechanisms, we aim to learn fine-grained representations for each timestamp without any compression along the time dimension. 
The output of the normalization layer is then fed to a linear embedding layer in the attention module, resulting in the projected vectors $\mathbf{H}_\text{emb} \in \mathbb{R}^{d}$.

% \vspace{-0.5em}
\begin{equation}
    % \small
    \mathbf{H}_\text{emb} = \mathbf{W}_{\text{emb}}\cdot\mathbf{H}
\end{equation}
% \vspace{-1.0em}

\noindent Here, $\mathbf{W}_{\text{emb}}$ is the weight matrix of the linear embedding layer.

Subsequently, self-attention operations are performed as follows.

% \vspace{-0.5em}
\begin{equation}
    % \small
    \label{eq:attention}
    \begin{aligned}
        \mathbf{Q} &= \mathbf{W}_{\mathbf{Q}} \cdot \mathbf{H}_{\text{emb}}\\
        \mathbf{K} &= \mathbf{W}_{\mathbf{K}} \cdot \mathbf{H}_{\text{emb}}\\
        \mathbf{V} &= \mathbf{W}_{\mathbf{V}} \cdot \mathbf{H}_{\text{emb}}\\
        \mathbf{S} &= \mathrm{softmax}\left(\frac{\mathbf{Q} \cdot \mathbf{K}^\top}{\sqrt{d}}\right)\\
        \mathbf{Y}_{1} &= \mathbf{S} \cdot \mathbf{V}
    \end{aligned}
\end{equation}
% \vspace{-0.5em}

\noindent Here, $\mathbf{W}_{\mathbf{Q}} \in \mathbb{R}^{d\times d}$, $\mathbf{W}_{\mathbf{K}} \in \mathbb{R}^{d\times d}$, and $\mathbf{W}_{\mathbf{V}} \in \mathbb{R}^{d\times d}$ are projection matrices for query, key, and value, respectively. 
In the specific implementation, we employ a multi-head self-attention mechanism, assuming a total of $M$ heads producing $M$ outputs $[\mathbf{Y}^{1}_{1},\ldots,\mathbf{Y}^{M}_{1}]$, where each attention head operates in a $\displaystyle\frac{d}{M}$ dimensional space.
Then, the outputs of the $M$ attention heads are concatenated and projected with a linear transformation, as shown in Equation~\ref{equation:multi-head_attention}.

% \vspace{-0.5em}
\begin{equation}
    % \small
    \label{equation:multi-head_attention}
    \mathbf{Y}_{1} = \mathbf{W}_{\text{mult}} \cdot [\mathbf{Y}^{1}_{1},\ldots, \mathbf{Y}^{M}_{1}]^\top
\end{equation}

\noindent Here, $\mathbf{W}_{\text{mult}}$ is a learnable parameter to conduct the linear transformation.

The output of multi-head self-attention is then fed to an addition and normalization layer to conduct a residual connection and normalization as shown in Equation~\ref{equation:add_norm}.

% \vspace{-0.5em}
\begin{equation}
    % \small
    \label{equation:add_norm}
    \mathbf{Y}_{2} = \mathbf{Y}_{1} + \frac{\mathbf{Y}_{1} - \mathbb{E}[\mathbf{Y}_{1}]}{\sqrt{\mathrm{Var}[\mathbf{Y}_{1}] + \epsilon}} \cdot \gamma_{2} + \beta_{2}
\end{equation}
% \vspace{-0.5em}

\noindent Here, the $\gamma_{2}$ and $\beta_{2}$ are learnable parameter vectors, $\mathbb{E}[\mathbf{Y}_{1}]$ and $\mathrm{Var}[\mathbf{Y}_{1}]$ are the expectation and variance of  matrix $\mathbf{Y}_{1}$.

The output of the addition and normalizing layer is fed to a multi-layer perceptron (MLP) to conduct a sequence of $k$ linear transformations. We use an  MLP with two linear layers and the ReLU activation function.

% \vspace{-0.5em}
\begin{equation}
    % \small
    \mathbf{Y}_{3} = \mathbf{W}_{2}\cdot\mathrm{ReLU}(\mathbf{W}_{1}\cdot\mathbf{Y}_{2})
\end{equation}
% \vspace{-1.0em}

\noindent Here, $\mathbf{W}_{1}$ and $\mathbf{W}_{2}$ are the learnable weight matrices of the MLP.
Finally, the output $\mathbf{Y}_{3}$ of the MLP is fed into the second addition and normalization layer, where the computation is similar to that in the first addition and normalization layer, to obtain $\mathbf{Y}$ (see Equation~\ref{equation:add_norm}).

\subsubsection{Stable and Auxiliary Feature Modules}
\label{subsubsec:stable-and-auxiliary-feature-modules}
The output of the attention module, $\mathbf{Y}$, is partitioned into two parts, and each part is fed into one of the two modules--the stable feature module and the auxiliary feature module. 
Stable features capture shared, invariant information across the time series, while auxiliary features capture local variations and noise.
Figure~\ref{fig:architecture}(c) shows these two modules.

More specifically, $\mathbf{Y} = [\mathbf{Y}_{\text{sta}}, \mathbf{Y}_{\text{aux}}]$, where $\displaystyle\mathbf{Y}_{\text{sta}} \in \mathbb{R}^{B\times\frac{d}{2}}$ and $\displaystyle\mathbf{Y}_{\text{aux}} \in \mathbb{R}^{B\times\frac{d}{2}}$ represent the separated representations that will be fed into the stable and auxiliary modules, respectively. 

To facilitate the model's effective learning of these two representations, we have designed both the stable feature module and the auxiliary feature module.
The stable features of a time series are the features that span many time steps to represent long-term patterns of the time series. 
In contrast, the auxiliary features of a time series are the features that only span a few time steps to represent short-term patterns or changes in individual observations of the time series.

To be able to distinguish between stable features and the auxiliary features, we first define two operations---a \textit{shuffle} operation and an \textit{identity} operation, which are used in both modules. 
The shuffle operation, $\mathrm{shuffle}(\cdot)$, performs random shuffling along the time dimension. The identity operation, $\mathrm{identity}(\cdot)$, represents no change to the input. The shuffle operation is applied along the time dimension within each subsequence window, while all feature dimensions are treated separately. This ensures that the temporal order of data points is manipulated, allowing us to distinguish between stable features (insensitive to shuffling) and auxiliary features (sensitive to shuffling). We proceed to elaborate on the auxiliary module and the stable module.

\noindent
\textbf{Auxiliary Module:} In the auxiliary module, we apply an identity operation to $\mathbf{Y}_{\text{sta}}$ and perform a shuffle operation on $\mathbf{Y}_{\text{aux}}$. Since the auxiliary features contain information related to specific timestamps, the shuffling of auxiliary features affects the final output sequence. Thus, we apply a shuffle operation to the input feature $\mathbf{Y}$, and the auxiliary features $\mathbf{Y}_{\text{aux}}$ to maintain consistency. By doing so, we aim to emphasize that the auxiliary features are strongly affected by shuffling because auxiliary features are associated with individual data points. When we modify the order of data points, we emphasize the prominent features of every single data point. Due to the complementary nature of stable features and auxiliary features, we concatenate the two types of features and project the concatenated feature space back to the original latent space. This way, $\mathbf{Y}_{\text{aux}}$ can capture rapidly changing features.

% \vspace{-0.5em}
\begin{equation}
    % \small
    \begin{aligned}
        \mathbf{Y}^I_{\text{sta}} &= \mathrm{identity}(\mathbf{Y}_{\text{sta}})\\ 
        \mathbf{Y}^S_{\text{aux}} &= \mathrm{shuffle}(\mathbf{Y}_{\text{aux}}) \\
        \mathbf{\hat{Y}}_{\text{aux}} &= \mathrm{concat}(\mathbf{Y}^I_{\text{sta}},\mathbf{Y}^S_{\text{aux}}) \cdot \mathbf{W}_{p}
    \end{aligned}
\end{equation}
% \vspace{-0.5em}

Then, we formulate the auxiliary loss that measures the closeness between the two representations, as follows. 

% \vspace{-0.5em}
\begin{equation}
% \small
    \mathcal{L}_{\text{aux}} = \|\mathrm{shuffle}(\mathbf{Y})- \mathbf{\hat{Y}}_{\text{aux}}\|_{\mathcal{F}}^{2}
    \label{equation:aux_loss}
\end{equation}
% \vspace{-0.5em}

\noindent
\textbf{Stable Module:} By definition, stable features remain relatively stable over a long period. Therefore a random perturbation of the stable features at a particular timestamp $i$, denoted as $\mathbf{Y}^S_{\text{sta}, i}$, should be interchangeable with its pre-perturbation stable feature  $\mathbf{Y}_{\text{sta}, i}$. Therefore, in the stable feature module, we only shuffle $\mathbf{Y}_{\text{sta}}$ while keeping $\mathbf{Y}_{\text{aux}}$ unchanged. By doing this, we aim to emphasize that the stable features are persistent and cannot be changed by shuffling because they are contained in long sequences. Finally, after concatenating these two types of features, we apply a projection to obtain the projected representation $\mathbf{\hat{Y}}_{\text{sta}}$.

% \vspace{-0.5em}
\begin{equation}
    % \small
    \begin{aligned}
        \mathbf{Y}^S_{\text{sta}} &= \mathrm{shuffle}(\mathbf{Y}_{\text{sta}})\\          
        \mathbf{Y}^I_{\text{aux}} &= \mathrm{identity}(\mathbf{Y}_{\text{aux}})\\
        \mathbf{\hat{Y}}_{\text{sta}} &= \mathrm{concat}(\mathbf{Y}^S_{\text{sta}},\mathbf{Y}^I_{\text{aux}}) \cdot \mathbf{W}_{p}
    \end{aligned}
\end{equation}
% \vspace{-0.5em}

%\subsubsection{Mutual information module}
The stable feature module lacks the self-supervisory information (i.e., shuffle) compared to the auxiliary module, {which considers the shuffled $\mathbf{Y}$  as self-supervisory information.} Using a loss function like the one used in the auxiliary module (Equation~\ref{equation:aux_loss}) makes it susceptible to learning a trivial solution that simply copies $\mathbf{Y}$, resulting in meaningless stable features. 
To avoid this issue, we follow the \textit{infomax principle}~\cite{DBLP:journals/neco/BellS95}, to maximize the mutual information between the input and output. Specifically, $\mathbf{Y}_{\text{sta}}$ contains the normal modes of $\mathbf{Y}$, so they should have a substantial amount of shared information. 
We then incorporate the training process of the mutual information estimator into the stable feature module to advocate maximizing the mutual information between $\mathbf{Y}$ and $\mathbf{Y}_{\text{sta}}$, denoted as $I_\theta(\mathbf{Y},\mathbf{Y}_{\text{sta}})$.
The final loss of the stable feature module is then defined as shown in Equation~\ref{equation:sta_loss}.
%} 

% \vspace{-0.5em}
\begin{equation}
    % \small
    \mathcal{L}_{\text{sta}} = \|\mathbf{Y}-\mathbf{\hat{Y}}_{\text{sta}}\|_{\mathcal{F}}^{2} - I_{\theta}(\mathbf{Y},\mathbf{{Y}}_{\text{sta}}) 
    \label{equation:sta_loss}
\end{equation}
% \vspace{-0.5em}

\noindent Here, $I_\theta(\cdot)$ is the mutual information estimator parameterized by $\theta$. 
We can choose a specific estimator among many existing ones. 
We choose $\mathrm{InfoNCE}$ as defined in Equation~\ref{equation:info_nce} as the default estimator due to its excellent performance as reported in recent studies~\cite{DBLP:conf/icml/BelghaziBROBHC18,DBLP:conf/iclr/HjelmFLGBTB19}. Later, we compare it empirically with other estimators (see Section~\ref{subsubsec:mutual_information_estimators}).

% \vspace{-0.5em}
\begin{equation}
    % \small
    I_{\mathrm{InfoNCE}} = \mathbb{E}_{\mathbb{P}(\mathbf{Y},\mathbf{{Y}}_{\text{sta}})}[f_{\theta}(\mathbf{Y},\mathbf{{Y}}_{\text{sta}})]-\mathbb{E}_{\mathbb{P}(\mathbf{{Y}}_{\text{sta}})}[\mathbb{E}_{\mathbb{P}(\mathbf{Y})}[e^{f_{\theta}(\mathbf{Y},\mathbf{{Y}}_{\text{sta}})}]] \label{equation:info_nce}
\end{equation}
% \vspace{-0.5em}

\noindent Here, $f_\theta(\mathbf{Y},\mathbf{{Y}}_{\text{sta}})$ is separable critic function defined as shown in Equation~\ref{eq:critic-function}.

% \vspace{-0.5em}
\begin{equation}
    % \small
    f_{\theta}(\mathbf{Y},\mathbf{{Y}}_{\text{sta}}) = \phi_{\theta}(\mathbf{Y})^\top\phi_{\theta}(\mathbf{{Y}}_{\text{sta}}),
    \label{eq:critic-function}
\end{equation}
% \vspace{-0.5em}

\noindent where $\phi_{\theta}(\cdot)$ is a non-linear transformation function such as a feed-forward neural network.

By proposing the loss function in Equation~\ref{equation:sta_loss}, we aim to achieve three targets. First, the MI measures the statistical dependency between latent representations and input data, offering greater robustness to the separation of the stable features from the original embedded features. Reconstruction error primarily captures point-wise deviations between input and output, which can be unreliable when anomalies are partially reconstructed--especially in the presence of contaminated training data. The MI can avoid the stable features converging into the contamination representation by also considering the number of observations. Because the number of anomalies is small, the anomalies even with large magnitudes cannot affect the MI severely. In this case, $\mathbf{Y}_{\text{sta}}$ encodes stable features of the time series, and short-term anomalies are less likely to distort this distribution. Second, this can be seen as introducing the maximization of MI into representation learning, a principle used widely~\cite{DBLP:conf/iclr/HjelmFLGBTB19,DBLP:journals/neco/BellS95}. This approach effectively prevents the model from learning trivial features. Third, it adds the critic function $f_{\theta}(\cdot)$ into the training process. This function is considered as a contrastive loss to provide self-supervisory information to the loss.

\subsection{Regularization}
\label{subsec:regularization}
\begin{figure}[t]
	\centering
	\includegraphics[width=1\linewidth]{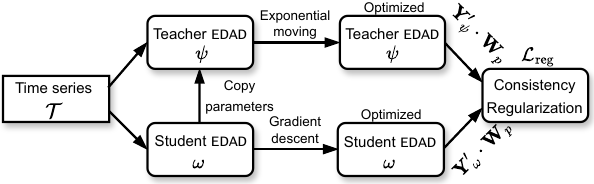}
        \caption{Regularization.} 
	\label{fig:regularization}
	% \vspace{-1.0em}
\end{figure}

We observe that both the stable and auxiliary feature modules are parameterized. Further, both modules are fed the output of the encoder, i.e., the stable features and auxiliary features. As a result, both modules share the parameters of the encoder for learning. However, these two types of features represent approximately orthogonal objectives, which can easily lead to conflicting parameter updates~\cite{DBLP:conf/iccv/PengZLWLL21}. We address this challenge by introducing a teacher–student architecture that serves as a form of consistency regularization~\cite{DBLP:conf/iclr/LaineA17}. This design encourages the shared encoder parameters to evolve more smoothly and coherently, despite the presence of competing learning signals. Figure~\ref{fig:regularization} illustrates the regularization process. Specifically, we make two copies of \texttt{EDAD}, where the decomposer is disabled and only the encoder is enabled, to serve as a teacher model and a student model. The student model is updated directly via gradient descent and is responsible for learning from the data at each training step~\cite{DBLP:journals/corr/Ruder16}. In contrast, the teacher model maintains an exponential moving average of the student's parameters~\cite{DBLP:conf/nips/TarvainenV17}. This design provides a smoother and more stable representation space that reduces the variance introduced by frequent student updates, thereby improving the reliability of mutual information estimation. We use $\omega$ and $\psi$ to represent the parameters of the student model and the teacher model, respectively. When computing the consistency regularization, we directly obtain the projected representation from the output representation of the encoder $\mathbf{Y}^{\prime}$ through the projection matrix $\mathbf{W}_{p}$. The consistency regularization for the student model and the teacher model is computed as shown in Equation~\ref{equation:reg_loss}. 

% \vspace{-0.5em}
\begin{equation}
    % \small
    \mathcal{L}_{\text{reg}} = \|\mathbf{Y}_{\omega}^{\prime}\cdot \mathbf{W}_p - \mathbf{Y}_{\psi}^{\prime}\cdot \mathbf{W}_p \|_{\mathcal{F}}^{2}
    \label{equation:reg_loss}
\end{equation}
% \vspace{-0.5em}

\noindent Here, $\mathbf{Y}_{\omega}^{\prime}$ represents the output representation of the student model, and $\mathbf{Y}_{\psi}^{\prime}$ represents the output representation of the teacher model.
This way, \texttt{EDAD} is enabled to utilize highly shared weights to partition the features of the time series into two parts, thereby increasing the robustness of the model training.

\subsection{Objective Function}

The overall loss is the weighted sum of the auxiliary reconstruction loss (Equation~\ref{equation:aux_loss}), the stable reconstruction loss (Equation~\ref{equation:sta_loss}), and the regularization (Equation~\ref{equation:reg_loss}).

% \vspace{-0.5em}
\begin{equation}
    % \small
    \mathcal{L} = \lambda_1\cdot\mathcal{L}_{\text{sta}}+\lambda_2 \cdot\mathcal{L}_{\text{aux}}+\lambda_3\cdot\mathcal{L}_{\text{reg}}
    \label{equation:objective_function}
\end{equation}
% \vspace{-0.5em}

\noindent Hyperparameters $\lambda_1$, $\lambda_2$, and $\lambda_3$  control the trade-off between the objective function terms. We  investigate the sensitivity to $\lambda_1$, $\lambda_2$, and $\lambda_3$ in the experimental study.

\subsection{Anomaly Scores}

We have shaped \texttt{EDAD} to enforce it on learning the stable features by augmenting the stable feature learning with the MI module. 
The remaining information related to individual observations and short-term patterns is maintained in the auxiliary features. 
Therefore, the shared information between the original features $\mathbf{Y}$ and the auxiliary features $\mathbf{Y}_\text{aux}$ can be used to identify anomalies, which are also related to individual observations and short-term patterns. This enables an ``asymmetric'' design of the loss function used for training and the definition of anomaly scores.

Given a time series subsequence, we can calculate an anomaly score as the point-wise mutual information between its encoded representation $\mathbf{Y}$ and the corresponding auxiliary representation $\mathbf{Y}_{\text{aux}}$. 

Due to the choice of different mutual information estimators, the critic function $f_{\theta}(\cdot)$ (see Equation~\ref{equation:sta_loss}) may not necessarily be proportional to $\small\displaystyle\frac{\mathbb{P}(\mathbf{Y}, \mathbf{{Y}}_{\text{aux}})}{\mathbb{P}(\mathbf{Y})\mathbb{P}(\mathbf{{Y}}_{\text{aux}})}$ when tightening the lower bound~\cite{DBLP:conf/iclr/TschannenDRGL20}, so it cannot be used alone to compute the anomaly scores. 
Therefore, we employ the entire estimator's forward pass. Let $I_{\theta}$ be the mutual information estimator parameterized by $\theta$. 
Then, we can compute the anomaly score for each data point $\mathbf{s}_{i}$ as follows. 

% \vspace{-0.5em}
\begin{equation}
% \small
   \mathcal{AS}(\mathbf{s}_{i}) = - I_{\theta}(\mathbf{Y}, \mathbf{{Y}}_{\text{aux}})   
\end{equation}
% \vspace{-0.5em}

A high score indicates that the input $\mathbf{Y}$ and $\mathbf{Y}_{\text{aux}}$ share less information. Since $\mathbf{Y}_{\text{aux}}$  
includes only short-term variations, $\mathbf{s}_{i}$ is more likely to be anomalous.

\section{Experiments}
\label{sec:exp}
\subsection{Experimental Settings}

\newcounter{dataset}
\setcounter{dataset}{1}
\newcounter{baseline}
\setcounter{baseline}{1}

\subsubsection{Datasets}
\label{subsubsec:datasets}
% \textcolor{red}{
We conduct experiments on eight real-world datasets that span a wide range of domains, such as manufacturing, natural sciences, and healthcare:
% (\rmnum{1}) Mars Science Laboratory (\textbf{MSL})~\cite{DBLP:conf/kdd/HundmanCLCS18}; 
(\thedataset\stepcounter{dataset}) Pooled Server Metrics (\textbf{{PSM}})~\cite{DBLP:conf/kdd/AbdulaalLL21} is collected from EBAY servers and records the server monitoring metrics; 
(\thedataset\stepcounter{dataset}) Soil Moisture Active Passive (\textbf{SMAP})~\cite{DBLP:conf/kdd/HundmanCLCS18} is collected by NASA and presents soil samples and telemetry information from the Mars exploration project; 
(\thedataset\stepcounter{dataset}) Secure Water Treatment (\textbf{SWAT})~\cite{DBLP:conf/cpsweek/MathurT16} is collected from a water treatment process in an infrastructure for research on cyber-security; 
(\thedataset\stepcounter{dataset}) Mars Science Laboratory (\textbf{MSL})~\cite{DBLP:conf/nips/LaiZXZWH21} is collected by NASA and shows the state of the sensors in the Mars exploration project;
% (\thedataset\stepcounter{dataset}) NIPSTS-GECCO (\textbf{GECCO})~\cite{DBLP:conf/nips/LaiZXZWH21} is collected from an Internet-of-things system and records drinking water quality;
(\thedataset\stepcounter{dataset}) NIPSTS-SWAN (\textbf{SWAN})~\cite{DBLP:conf/nips/LaiZXZWH21} is extracted from solar photospheric vector magnetograms in Spaceweather HMI Active Region Patch series;  
(\thedataset\stepcounter{dataset}) \textbf{KDD21}~\cite{DBLP:journals/pvldb/PaparrizosKBTPF22} is a composite dataset released for a SIGKDD 2021 competition;
(\thedataset\stepcounter{dataset}) Numenta Anomaly Benchmark (\textbf{NAB})~\cite{DBLP:journals/ijon/AhmadLPA17} comprises labeled time series data from diverse sources, encompassing AWS server metrics, online ad click rates, real-time traffic data, and Twitter mentions of major publicly traded firms;
(\thedataset\stepcounter{dataset}) Supraventricular Arrhythmia Database (\textbf{SVDB})~\cite{Moody_Mark_2001} includes 78 half-hour ECG recordings that supplement supraventricular arrhythmias in the MIT-BIH Arrhythmia Database. 
The eight datasets encompass both multivariate and univariate time series. We acknowledge that datasets such as \textbf{SWaT}, \textbf{SMAP}, and \textbf{MSL} have known limitations, including high anomaly density, inconsistent labels, long anomaly windows, and unrealistic distributions. These issues are discussed in \texttt{TimeSeAD}~\cite{DBLP:journals/tmlr/WagnerMSNRK23}. Nevertheless, these datasets are widely used in the time-series anomaly detection literature, which motivated our decision to include them in our experiments. We provide statistical information on the experimental datasets in Table~\ref{table:dataset_description}, including the dimensionality of each dataset, its length, and the proportion of anomalies.
% }

\subsubsection{Baselines}  
We compare \framework with thirteen strong and well-known anomaly detection methods. To be comprehensive, we include neural network based anomaly detection methods as well as traditional anomaly detection methods with good performance and published in top venues. Specifically, we include eleven methods:
(\thebaseline\stepcounter{baseline}) \texttt{OC-SVM}~\cite{DBLP:journals/ml/TaxD04} learns a boundary that encompasses the normal data while leaving anomalies outside the boundary;
(\thebaseline\stepcounter{baseline}) \texttt{IForest}~\cite{DBLP:conf/icdm/LiuTZ08} uses an ensemble of isolation trees to detect anomalies;
(\thebaseline\stepcounter{baseline}) \texttt{DAGMM}~\cite{DBLP:conf/iclr/ZongSMCLCC18} integrates a \texttt{GMM} and \texttt{AE} to model the distribution of multidimensional data;
% 
% (\thebaseline\stepcounter{baseline}) \texttt{MatrixProfile}~\cite{DBLP:conf/icdm/YehZUBDDSMK16} {is an anomaly detection} algorithm based on time series subsequence similarity calculation;
%
(\thebaseline\stepcounter{baseline}) \texttt{Series2Graph}~\cite{boniol2020series2graph} is an anomaly detection algorithm that transforms time series into graph structures;
(\thebaseline\stepcounter{baseline}) \texttt{SAND}~\cite{boniol2021sand} is an anomaly detection algorithm designed for streaming data. It identifies anomalous patterns by clustering input data sequences;
(\thebaseline\stepcounter{baseline}) \texttt{LSTM-AD}~\cite{DBLP:conf/kdd/HundmanCLCS18} uses \texttt{RNN}s to detect anomalies by forecasting over long sequences of data;
(\thebaseline\stepcounter{baseline}) \texttt{MAD-GAN}~\cite{DBLP:conf/icann/LiCJSGN19} employs \texttt{GAN} to recognize anomalies by reconstructing testing samples from the latent space;
(\thebaseline\stepcounter{baseline}) \texttt{TranAD}~\cite{Tuli_Casale_Jennings} utilizes transformer models to infer anomalies by considering broader temporal trends in the data; 
(\thebaseline\stepcounter{baseline}) \texttt{GDN}~\cite{Deng_Hooi_2022} integrates GNNs and meta-learning with past and recent information to enable anomaly detection;
(\thebaseline\stepcounter{baseline}) \texttt{OmniAnomaly}~\cite{DBLP:conf/kdd/SuZNLSP19} integrates \texttt{GRU}s and \texttt{VAE}s to learn robust representations of time series data;
(\thebaseline\stepcounter{baseline}) \texttt{IMdiffusion}~\cite{imdiffusion} combines time series imputation and diffusion models to achieve robust anomaly detection.
(\thebaseline\stepcounter{baseline}) \texttt{AnomalyTrans}~\cite{DBLP:conf/iclr/XuWWL22} models prior associations and series associations to capture the association discrepancies;
(\thebaseline\stepcounter{baseline}) \texttt{DCdetector}~\cite{DBLP:conf/kdd/YangZZW023} detects time series anomalies using robust representations based on contrastive learning.
Note that we use the publicly available implementations from the authors of the above methods.
% Details of these baselines are provided in Appendix~\ref{ap:baselines}.
% }

\subsubsection{Metrics}
We use standard metrics for anomaly detection, including Precision (\textit{P}), Recall (\textit{R}), F1-score (\textit{F1}), Area Under the Precision-Recall Curve (\textit{A-PR}), and Area Under the Receiver Operating Characteristic Curve (\textit{A-ROC})~\cite{DBLP:conf/kdd/LiZHSJWP21}. In addition, we report
Volume-under-the-Surface of Precision-Recall (\textit{V-PR}) and Volume-under-the-Surface of Receiver Operating Characteristic (\textit{V-ROC})~\cite{vpr} to alleviate bias stemming from threshold selections and provide an alternative evaluation perspective on anomaly detection methods, utilizing continuous buffer regions~\cite{DBLP:journals/pvldb/PaparrizosKBTPF22}. Each metric offers valuable information. 
% It is worth noting that EDAD achieves the highest F1 by a large margin, which proves that EDAD can identify potentially hard anomalies compared to other baselines. 

 % i think we should emphasis that EDAD achieve best F1 by a large margin，demonstrate EDAD is able to excavate  few and hard anomalies in GECCO
 
% \textcolor{red}{What are the differences between A-PR and V-PR, between A-ROC and V-ROC?}

\subsubsection{Implementation Details}
\label{subsubsec:implementation}
We implement the proposed framework and baselines by utilizing PyTorch~\cite{DBLP:conf/nips/PaszkeGMLBCKLGA19} and Scikit-learn 0.24~\cite{DBLP:journals/jmlr/PedregosaVGMTGBPWDVPCBPD11} in Python 3.10. All experiments were executed on a cluster server, which runs Linux Ubuntu 18.04.6 LTS. The server is equipped with an NVIDIA Tesla-A800 GPU with two 64-core AMD CPUs and 512 GiB RAM. The source code is available at \underline{\url{https://github.com/zhangbububu/EDAD/}}.

\begin{table}[t]
	\footnotesize
    % \fontsize{7.5pt}{7.5pt}\selectfont
    % \small
    \tabcolsep3pt
	\caption{Dataset statistics.}
    \label{table:datasets}
	\centering
	{
		\begin{tabular}{c|lll}
			\toprule
			\textbf{Dataset} & \textbf{Dimension} & \textbf{Average Length} & \textbf{Anomaly Ratio (\%)} \\
			\midrule
			% \textbf{MSL}     & 55                 & 132,046                 & 10.5             \\
                \textbf{PSM}     & 25                 & 220,322                 & 27.8             \\
			\textbf{SMAP}    & 25                 & 562,800                 & 12.8             \\
			\textbf{SWAT}    & 51                 & 944,919                 & 12.0             \\
                \textbf{MSL
                }   & 55                  & 132,046                 & 10.5              \\
			\textbf{SWAN}    & 38                 & 120,000                 & 32.6             \\
                \textbf{KDD21}   & 1                  & 77,415                  & 10.67            \\
			\textbf{NAB}     & 1                  & 6,301                   & 2.67             \\
			\textbf{SVDB}    & 1                  & 230,400                 & 4.68             \\
			\bottomrule
		\end{tabular}
	}
	\label{table:dataset_description}
	% \vspace{-2.0em}
\end{table}

\subsubsection{Hyperparameter Settings}
\label{subsubsec:hyperparameter}
Following recent studies~\cite{DBLP:conf/kdd/YangZZW023,Tuli_Casale_Jennings,DBLP:conf/iclr/XuWWL22}, the dimensionality $d$ of the hidden layer is set to 256, the number of encoder layers is set to 3, the number of heads $M$ in the multi-head attention is set to 8, and the window of the input model $B$ is set to 100. By doing this we ensure a similar backbone and the fairness of comparison. 
We set the anomaly ratio to 1\% so that the 1\% of the data points with the highest anomaly scores are anomalies \cite{DBLP:conf/iclr/XuWWL22}.
We use the \texttt{InfoNCE}~\cite{DBLP:journals/corr/abs-1807-03748} with separable critics as the mutual information estimator.
In addition, we use the Adam optimizer~\cite{DBLP:journals/corr/KingmaB14} with a learning rate of $5\times10^{-4}$ for model training. 
Early stopping is adopted in the training process. 

To tune $\lambda_{1}$, $\lambda_{2}$, and $\lambda_{3}$, we vary each of $\lambda_{1}$, $\lambda_{2}$, and $\lambda_{3}$ among 0.1, 0.5, 1, 2, and 3. After getting the results for all combinations of $\lambda_{1}$, $\lambda_{2}$, and $\lambda_{3}$, we identify the median result and use the corresponding hyperparameter setting as the default setting. We do not use the best result because, in unsupervised settings, we have no labeled data to enable identifying the best result. Further, we conduct experiments to study the sensitivity of different $\lambda_{1}$, $\lambda_{2}$, and $\lambda_{3}$ in Section~\ref{subsubsec:lambda_effect}. To do so, we vary a chosen hyperparameter in its range while fixing the other hyperparameters to their default values. 
We also study the effect of window size $B$ in Section~\ref{subsubsec:win_size_effect}.

For the other baselines, we use the hyperparameter settings recommended in existing studies if provided. Otherwise, we randomly vary parameters in specific methods, such as the kernel degree in \texttt{OC-SVM}. Then, we report the median of multiple runs using different hyperparameters.

\subsection{Experimental Results}

\begin{table*}[htbp]
\centering
\fontsize{7.0pt}{7.0pt}\selectfont
\tabcolsep2.5pt
\caption{\textit{P}, \textit{R}, \textit{F1}, \textit{A-PR}, \textit{A-ROC}, \textit{V-PR}, and \textit{V-ROC} of anomaly detection methods. The top three highest accuracies are highlighted in blue, where the best and the runner-up results are in \textbf{bold} and \underline{underlined} text, respectively.}
\label{table:main_result}
% \resizebox{\textwidth}{!}{
\begin{tabular}{l|ccccccc|ccccccc|ccccccc}
\hline
\multirow{2}{*}{\textbf{Method}} & \multicolumn{7}{c|}{\textbf{PSM}} & \multicolumn{7}{c|}{\textbf{SMAP}} & \multicolumn{7}{c}{\textbf{SWAT}} \\
\cline{2-22}
& \textit{P}               & \textit{R}               & \textit{F1}       & \textit{A-PR} &    \textit{A-ROC}   & \textit{V-PR}  & \textit{V-ROC} & \textit{P}               & \textit{R}               & \textit{F1}       & \textit{A-PR} &    \textit{A-ROC}   & \textit{V-PR}  & \textit{V-ROC} & \textit{P}               & \textit{R}               & \textit{F1}       & \textit{A-PR} &    \textit{A-ROC}   & \textit{V-PR}  & \textit{V-ROC} \\
\hline

%%%%
\texttt{OC-SVM} & 0.627 & 0.706 & 0.664 & 0.417 & 0.619 & 0.369 & 0.531 & 0.512 & 0.578 & 0.543 & 0.101 & 0.392 & 0.113 & \cell{\underline{0.518}} & 0.419 & 0.478 & 0.447 & 0.126 & \cell{\underline{0.657}} & 0.133 & 0.477 \\
\texttt{IForest} & 0.627 & 0.924 & 0.834 & 0.334 & 0.542 & 0.334 & 0.541 & 0.523 & 0.590 & 0.555 & 0.121 & 0.487 & 0.135 & 0.499 & 0.492 & 0.449 & 0.470 & 0.093 & 0.345 & 0.129 & 0.424 \\
\texttt{DAGMM} & 0.934 & 0.700 & 0.801 & 0.430 & 0.647 & 0.354 & 0.515 & 0.864 & 0.567 & 0.685 & 0.135 & 0.561 & 0.123 & 0.468 & 0.861 & 0.530 & 0.656 & \cell{\textbf{0.207}} & \cell{\textbf{0.710}} & \cell{0.241} & \cell{\textbf{0.538}} \\
% \texttt{MatrixProfile} & 0.862 & 0.879 & 0.870 & 0.433 & 0.489 & 0.327 & 0.505 & 0.914 & 0.786 & 0.845 & 0.133 & 0.578 & 0.135 & 0.467 & 0.853 & 0.711 & 0.776 & \cell{\underline{0.170}} & 0.384 & 0.172 & 0.443 \\
\texttt{Series2Graph} & 0.906 & 0.893 & 0.899 & \cell{\textbf{0.546}} & 0.471 & 0.313 & 0.512 & 0.903 & 0.689 & 0.782 & 0.114 & \cell{0.584} & 0.137 & 0.492 & 0.855 & 0.809 & 0.831 & \cell{0.161} & 0.280 & \cell{\underline{0.247}} & 0.392 \\
\texttt{SAND} & 0.931 & 0.861 & 0.895 & 0.415 & 0.479 & 0.401 & 0.542 & 0.927 & 0.826 & 0.874 & \cell{\underline{0.154}} & 0.455 & 0.146 & 0.502 & 0.867 & 0.713 & 0.782 & 0.142 & 0.343 & 0.179 & 0.463 \\
\texttt{LSTM-AD} & 0.769 & 0.896 & 0.828 & \cell{\underline{0.537}} & \cell{\textbf{0.714}} & \cell{\textbf{0.523}} & 0.526 & 0.894 & 0.781 & 0.833 & 0.142 & 0.579 & 0.122 & 0.458 & 0.861 & 0.832 & 0.846 & 0.094 & 0.405 & 0.101 & 0.250 \\
\texttt{MAD-GAN} & \cell{\textbf{0.986}} & 0.772 & 0.866 & \cell{0.524} & \cell{\underline{0.687}} & \cell{\underline{0.451}} & \cell{\textbf{0.601}} & 0.678 & 0.603 & 0.638 & 0.103 & 0.423 & 0.118 & 0.459 & 0.791 & 0.542 & 0.643 & 0.139 & 0.317 & 0.113 & 0.350 \\
\texttt{TranAD} & 0.950 & 0.895 & 0.922 & 0.511 & 0.665 & 0.352 & \cell{\underline{0.571}} & 0.822 & 0.850 & 0.836 & 0.113 & 0.416 & \cell{\underline{0.156}} & 0.425 & 0.702 & 0.726 & 0.714 & 0.126 & 0.323 & 0.235 & 0.356 \\
\texttt{GDN} & 0.875 & 0.838 & 0.856 & 0.438 & 0.657 & 0.355 & 0.475 & 0.907 & 0.612 & 0.731 & 0.096 & 0.375 & 0.112 & 0.414 & 0.171 & 0.058 & 0.086 & 0.119 & 0.312 & 0.113 & 0.351 \\
\texttt{OmniAnomaly} & 0.883 & 0.744 & 0.808 & 0.419 & 0.627 & \cell{0.439} & 0.522 & 0.924 & 0.819 & 0.869 & 0.097 & 0.378 & 0.113 & 0.417 & 0.814 & 0.843 & 0.828 & 0.121 & 0.338 & 0.113 & 0.351 \\
\texttt{IMdiffusion} & \cell{0.975} & 0.875 & 0.923 & 0.345 & 0.569 & 0.337 & 0.545 & 0.923 & 0.889 & 0.906 & 0.113 & 0.468 & 0.131 & 0.506 & \cell{\underline{0.932}} & 0.876 & 0.903 & 0.129 & 0.544 & 0.157 & 0.503 \\
\texttt{AnomalyTrans} & 0.969 & \cell{0.978} & \cell{0.973} & 0.396 & 0.298 & 0.277 & 0.486 & \cell{0.935} & \cell{\textbf{0.994}} & \cell{0.964} & \cell{\textbf{0.171}} & \cell{\underline{0.595}} & \cell{\textbf{0.157}} & \cell{0.509} & 0.891 & \cell{0.992} & \cell{0.939} & 0.071 & 0.179 & 0.109 & 0.434 \\
\texttt{DCdetector} & 0.973 & \cell{\textbf{0.985}} & \cell{\underline{0.979}} & 0.462 & 0.481 & 0.276 & 0.490 & \cell{\underline{0.955}} & \cell{\underline{0.988}} & \cell{\underline{0.970}} & \cell{0.151} & 0.580 & 0.147 & 0.502 & \cell{\underline{0.932}} & \cell{\underline{0.996}} & \cell{\underline{0.963}} & 0.157 & \cell{0.604} & 0.149 & \cell{0.507} \\ \hline
\texttt{EDAD} (\textbf{ours}) & \cell{\underline{0.978}} & \cell{\underline{0.984}} & \cell{\textbf{0.981}} & 0.517 & \cell{0.669} & 0.382 & \cell{0.549} & \cell{\textbf{0.970}} & \cell{0.974} & \cell{\textbf{0.972}} & 0.147 & \cell{\textbf{0.599}} & \cell{0.149} & \cell{\textbf{0.535}} & \cell{\textbf{0.938}} & \cell{\textbf{1.000}} & \cell{\textbf{0.968}} & \cell{0.172} & 0.571 & \cell{\textbf{0.334}} & \cell{\underline{0.512}} \\ 
\hline

%%%%%%%%%%%%%%

\multirow{2}{*}{\textbf{Method}} & \multicolumn{7}{c|}{\textbf{MSL}} & \multicolumn{7}{c|}{\textbf{SWAN}} & \multicolumn{7}{c}{\textbf{KDD21}} \\
\cline{2-22}
& \textit{P}               & \textit{R}               & \textit{F1}       & \textit{A-PR} &    \textit{A-ROC}   & \textit{V-PR}  & \textit{V-ROC} & \textit{P}               & \textit{R}               & \textit{F1}       & \textit{A-PR} &    \textit{A-ROC}   & \textit{V-PR}  & \textit{V-ROC} & \textit{P}               & \textit{R}               & \textit{F1}       & \textit{A-PR} &    \textit{A-ROC}   & \textit{V-PR}  & \textit{V-ROC} \\
\hline

\texttt{OC-SVM} & 0.602 & 0.873 & 0.713 & 0.185 & 0.593 & 0.207 & 0.663 & 0.474 & 0.498 & 0.486 & 0.326 & \cell{\underline{0.501}} & 0.318 & \cell{\underline{0.509}} & 0.173 & 0.625 & 0.271 & \cell{0.022} & 0.502 & 0.025 & \cell{\textbf{0.657}} \\
\texttt{IForest} & 0.541 & 0.863 & 0.665 & 0.173 & 0.570 & 0.191 & 0.649 & 0.570 & \cell{\underline{0.598}} & 0.583 & 0.379 & 0.487 & 0.375 & 0.440 & \cell{\underline{0.309}} & 0.607 & \cell{0.410} & \cell{\textbf{0.042}} & 0.561 & \cell{\textbf{0.039}} & 0.631 \\
\texttt{DAGMM} & 0.894 & 0.637 & 0.744 & 0.159 & 0.566 & 0.171 & 0.650 & 0.436 & 0.391 & 0.412 & 0.471 & 0.472 & 0.349 & 0.403 & 0.213 & 0.558 & 0.308 & 0.015 & \cell{\underline{0.698}} & 0.032 & 0.621 \\
% \texttt{MatrixProfile} & 0.752 & 0.866 & 0.805 & 0.180 & 0.560 & 0.189 & 0.654 & 0.801 & 0.522 & 0.632 & 0.459 & 0.414 & 0.373 & 0.435 & 0.233 & 0.575 & 0.332 & \cell{0.025} & 0.505 & 0.031 & 0.618 \\
\texttt{Series2Graph} & \cell{\textbf{0.937}} & 0.898 & \cell{0.917} & 0.176 & 0.533 & 0.193 & 0.608 & 0.745 & \cell{\textbf{0.609}} & 0.670 & 0.401 & 0.381 & 0.343 & 0.467 & 0.151 & 0.593 & 0.241 & 0.018 & 0.484 & 0.030 & 0.622 \\
\texttt{SAND} & 0.875 & 0.817 & 0.845 & \cell{0.194} & 0.569 & 0.188 & 0.656 & 0.837 & 0.575 & 0.682 & 0.396 & 0.370 & 0.393 & 0.478 & 0.218 & \cell{0.642} & 0.325 & 0.016 & 0.499 & \cell{0.033} & 0.623 \\
\texttt{LSTM-AD} & 0.858 & 0.828 & 0.842 & 0.188 & \cell{\underline{0.616}} & \cell{0.214} & \cell{\textbf{0.693}} & 0.474 & 0.211 & 0.292 & 0.454 & 0.463 & 0.329 & 0.471 & 0.215 & 0.550 & 0.309 & 0.013 & 0.444 & 0.030 & 0.618 \\
\texttt{MAD-GAN} & 0.723 & 0.772 & 0.746 & 0.190 & 0.599 & 0.209 & 0.675 & 0.921 & 0.589 & 0.718 & \cell{\textbf{0.495}} & \cell{\underline{0.501}} & 0.422 & 0.478 & 0.100 & 0.615 & 0.172 & 0.019 & 0.264 & 0.029 & \cell{0.634} \\
\texttt{TranAD} & 0.890 & \cell{0.931} & 0.910 & 0.193 & 0.515 & \cell{\textbf{0.217}} & 0.578 & \cell{0.939} & 0.579 & 0.716 & \cell{0.477} & 0.499 & 0.311 & 0.382 & 0.097 & 0.595 & 0.167 & \cell{\underline{0.036}} & 0.327 & 0.030 & 0.623 \\
\texttt{GDN} & \cell{\underline{0.933}} & 0.687 & 0.791 & 0.191 & \cell{0.603} & 0.211 & 0.674 & 0.928 & 0.528 & \cell{\underline{0.735}} & \cell{\underline{0.485}} & 0.474 & 0.424 & 0.478 & 0.102 & 0.615 & 0.175 & 0.015 & 0.276 & 0.028 & 0.631 \\
\texttt{OmniAnomaly} & 0.886 & 0.859 & 0.872 & 0.189 & 0.601 & 0.213 & 0.679 & 0.834 & 0.461 & 0.594 & 0.472 & \cell{\textbf{0.503}} & \cell{\textbf{0.454}} & 0.456 & 0.102 & 0.619 & 0.175 & 0.018 & 0.690 & 0.029 & \cell{\underline{0.641}} \\
\texttt{IMdiffusion} & 0.919 & \cell{\textbf{0.961}} & \cell{\underline{0.940}} & 0.160 & 0.531 & 0.179 & 0.594 & 0.932 & 0.566 & 0.597 & 0.292 & 0.481 & \cell{0.429} & 0.477 & 0.290 & 0.591 & 0.389 & 0.015 & 0.655 & 0.031 & 0.611 \\
\texttt{AnomalyTrans} & 0.930 & 0.893 & 0.911 & \cell{\textbf{0.215}} & 0.583 & \cell{\underline{0.216}} & \cell{0.682} & 0.907 & 0.474 & 0.622 & 0.222 & 0.257 & 0.402 & 0.490 & 0.097 & 0.595 & 0.167 & 0.010 & 0.578 & 0.020 & 0.619 \\
\texttt{DCdetector} & 0.892 & 0.867 & 0.879 & 0.193 & 0.572 & 0.206 & \cell{\underline{0.683}} & \cell{\underline{0.951}} & \cell{0.595} & \cell{0.732} & 0.286 & 0.411 & 0.346 & \cell{0.494} & \cell{0.304} & \cell{\underline{0.708}} & \cell{\underline{0.425}} & 0.015 & \cell{\textbf{0.723}} & 0.023 & 0.616 \\ \hline
\texttt{EDAD} (\textbf{ours}) & \cell{0.931} & \cell{\textbf{0.961}} & \cell{\textbf{0.946}} & \cell{\underline{0.197}} & \cell{\textbf{0.619}} & 0.194 & 0.677 & \cell{\textbf{0.980}} & 0.593 & \cell{\textbf{0.739}} & 0.326 & \cell{\underline{0.501}} & \cell{\underline{0.434}} & \cell{\textbf{0.512}} & \cell{\textbf{0.310}} & \cell{\textbf{0.740}} & \cell{\textbf{0.437}} & 0.017 & \cell{0.693} & \cell{\underline{0.035}} & 0.633 \\
\hline

%%%%%%%%

\multirow{2}{*}{\textbf{Method}} & \multicolumn{7}{c|}{\textbf{NAB}} & \multicolumn{7}{c|}{\textbf{SVDB}} & \multicolumn{7}{c}{\textbf{AVERAGE}} \\
\cline{2-22}
& \textit{P}               & \textit{R}               & \textit{F1}       & \textit{A-PR} &    \textit{A-ROC}   & \textit{V-PR}  & \textit{V-ROC} & \textit{P}               & \textit{R}               & \textit{F1}       & \textit{A-PR} &    \textit{A-ROC}   & \textit{V-PR}  & \textit{V-ROC} & \textit{P}               & \textit{R}               & \textit{F1}       & \textit{A-PR} &    \textit{A-ROC}   & \textit{V-PR}  & \textit{V-ROC} \\
\hline

\texttt{OC-SVM} & 0.437 & \cell{0.983} & 0.605 & \cell{\underline{0.336}} & 0.483 & \cell{\underline{0.311}} & 0.614 & 0.462 & \cell{\textbf{0.986}} & 0.629 & \cell{\textbf{0.239}} & 0.526 & \cell{\underline{0.230}} & \cell{\underline{0.663}} & 0.463 & 0.716 & 0.545 & 0.219 & 0.534 & 0.213 & \cell{\underline{0.579}} \\
\texttt{IForest} & 0.765 & 0.774 & 0.769 & 0.146 & 0.434 & 0.230 & 0.627 & \cell{0.812} & 0.732 & 0.770 & 0.171 & \cell{\textbf{0.594}} & 0.212 & 0.651 & 0.580 & 0.692 & 0.632 & 0.182 & 0.503 & 0.206 & 0.558 \\
\texttt{DAGMM} & 0.501 & 0.589 & 0.541 & \cell{\textbf{0.360}} & 0.450 & \cell{0.304} & \cell{\underline{0.647}} & 0.621 & 0.635 & 0.628 & 0.104 & 0.360 & 0.215 & 0.661 & 0.666 & 0.576 & 0.597 & \cell{\textbf{0.235}} & \cell{0.558} & 0.224 & 0.563 \\
% \texttt{MatrixProfile} & 0.690 & 0.892 & 0.778 & 0.176 & 0.416 & 0.281 & \cell{\textbf{0.648}} & \cell{\underline{0.817}} & 0.928 & 0.753 & 0.161 & 0.323 & 0.219 & 0.639 & 0.717 & 0.770 & 0.724 & 0.217 & 0.459 & 0.216 & 0.551 \\
\texttt{Series2Graph} & 0.798 & 0.829 & 0.813 & 0.218 & 0.547 & \cell{\textbf{0.326}} & 0.625 & 0.745 & 0.917 & 0.822 & 0.163 & 0.327 & 0.205 & 0.625 & 0.755 & 0.780 & 0.747 & 0.225 & 0.451 & 0.224 & 0.543 \\
\texttt{SAND} & 0.730 & 0.900 & 0.806 & 0.249 & 0.456 & 0.251 & \cell{0.639} & 0.755 & 0.847 & 0.798 & 0.187 & 0.427 & 0.226 & \cell{0.662} & 0.768 & 0.773 & 0.751 & 0.219 & 0.450 & \cell{0.227} & \cell{0.571} \\
\texttt{LSTM-AD} & 0.733 & 0.821 & 0.775 & 0.242 & 0.411 & 0.261 & \cell{0.638} & 0.803 & 0.877 & \cell{0.838} & 0.144 & 0.456 & \cell{\underline{0.230}} & 0.658 & 0.701 & 0.725 & 0.695 & \cell{0.227} & 0.511 & 0.226 & 0.539 \\
\texttt{MAD-GAN} & 0.736 & 0.898 & 0.809 & 0.165 & 0.317 & 0.246 & 0.632 & 0.619 & 0.924 & 0.741 & 0.117 & 0.282 & 0.227 & 0.657 & 0.694 & 0.714 & 0.667 & 0.219 & 0.424 & \cell{0.227} & 0.561 \\
\texttt{TranAD} & 0.743 & 0.920 & 0.822 & 0.123 & 0.587 & 0.246 & 0.629 & 0.610 & 0.884 & 0.722 & 0.105 & 0.508 & 0.223 & 0.632 & 0.719 & 0.798 & 0.726 & 0.211 & 0.480 & 0.221 & 0.525 \\
\texttt{GDN} & 0.753 & 0.928 & 0.831 & 0.115 & \cell{0.651} & 0.248 & 0.632 & 0.618 & 0.923 & 0.740 & 0.198 & \cell{\underline{0.575}} & 0.225 & 0.654 & 0.661 & 0.649 & 0.618 & 0.207 & 0.490 & 0.215 & 0.539 \\
\texttt{OmniAnomaly} & 0.740 & 0.920 & 0.820 & 0.213 & \cell{\underline{0.652}} & 0.243 & 0.633 & 0.625 & \cell{\underline{0.938}} & 0.750 & 0.162 & 0.284 & 0.227 & 0.657 & 0.726 & 0.775 & 0.715 & 0.211 & 0.509 & \cell{\underline{0.229}} & 0.545 \\
\texttt{IMdiffusion} & \cell{\underline{0.915}} & 0.846 & \cell{0.879} & 0.260 & 0.638 & 0.245 & 0.631 & 0.719 & 0.924 & 0.809 & 0.217 & 0.415 & 0.193 & 0.624 & \cell{0.826} & 0.816 & \cell{0.793} & 0.191 & 0.538 & 0.213 & 0.561 \\
\texttt{AnomalyTrans} & 0.743 & 0.920 & 0.822 & 0.227 & 0.302 & 0.219 & 0.615 & \cell{0.811} & 0.865 & 0.837 & \cell{0.225} & 0.320 & 0.197 & 0.571 & 0.785 & \cell{0.839} & 0.779 & 0.192 & 0.389 & 0.200 & 0.551 \\
\texttt{DCdetector} & \cell{\underline{0.915}} & \cell{\underline{0.996}} & \cell{\underline{0.954}} & 0.228 & 0.605 & 0.207 & 0.616 & 0.633 & 0.892 & \cell{\underline{0.853}} & 0.213 & \cell{0.550} & 0.190 & 0.563 & \cell{\underline{0.842}} & \cell{\underline{0.878}} & \cell{\underline{0.844}} & 0.213 & \cell{\underline{0.566}} & 0.193 & 0.559 \\ \hline
\texttt{EDAD} (\textbf{ours}) & \cell{\textbf{0.919}} & \cell{\textbf{0.997}} & \cell{\textbf{0.956}} & \cell{0.262} & \cell{\textbf{0.661}} & 0.290 & 0.636 & \cell{\textbf{0.828}} & \cell{0.933} & \cell{\textbf{0.877}} & \cell{\underline{0.231}} & 0.532 & \cell{\textbf{0.248}} & \cell{\textbf{0.668}} & \cell{\textbf{0.857}} & \cell{\textbf{0.898}} & \cell{\textbf{0.860}} & \cell{\underline{0.232}} & \cell{\textbf{0.606}} & \cell{\textbf{0.258}} & \cell{\textbf{0.590}} \\
\hline

\end{tabular}
% } % resize box here
% \vspace{-1.0em}
\end{table*}

\begin{table}[h]
\footnotesize
\caption{Overall ranking of anomaly detection methods.}
\label{table:ranking}
\centering
    \begin{tabular}{l|ccc}
        \toprule
        \textbf{Method} & \textbf{1st} & \textbf{2nd} & \textbf{3rd}  \\ 
        \midrule
        \texttt{OC-SVM} & 3 & 8 & 3 \\
        \texttt{IForest} & 3 & 2 & 2 \\
        \texttt{DAGMM} & 4 & 2 & 2 \\
        \texttt{Series2Graph} & 4 & 1 & 3 \\
        \texttt{SAND} & 0 & 1 & 5 \\
        \texttt{LSTM-AD} & 3 & 3 & 3 \\
        \texttt{MAD-GAN} & 3 & 3 & 3 \\
        \texttt{TranAD} & 1 & 3 & 3 \\
        \texttt{GDN} & 0 & 4 & 2 \\
        \texttt{OmniAnomaly} & 2 & 3 & 1 \\
        \texttt{IMdiffusion} & 1 & 4 & 5 \\
        \texttt{AnomalyTrans} & 4 & 2 & 9 \\
        \texttt{DCdetector} & 2 & 15 & 10 \\ \hline
        \texttt{EDAD} (\textbf{ours}) & 26 & 9 & 10 \\
        \bottomrule
    \end{tabular}
    % \vspace{-15pt}
\end{table}

\begin{table}[!ht]
% \vspace{-1.0em}
\fontsize{7.5pt}{7.5pt}\selectfont
\tabcolsep1.0pt
\centering
\caption{\textit{P}, \textit{R}, \textit{F1}, \textit{A-PR}, \textit{A-ROC}, \textit{V-PR}, and \textit{V-ROC}  of variants of \texttt{EDAD} averaged over the nine  datasets. \textcolor{black}{The second block represents the estimator, and the third block represents the critic function. The symbol $\circ$ indicates that we use the corresponding estimator/critic function instead of the default one. The top three highest accuracies are highlighted with blue, where the best and the runner-up results are in \textbf{bold} and \underline{underline} text, respectively.}
}

\label{table:main_result_v2_ablation_avg}
% \vspace{-1.em}
% \resizebox{\columnwidth}{!}{
\begin{tabular}{l|P{0.8cm}P{0.8cm}P{0.8cm}P{0.8cm}P{0.8cm}P{0.8cm}P{0.8cm}}
\hline
\textbf{Method}               & \textit{P}               & \textit{R}               & \textit{F1}       & \textit{A-PR} &    \textit{A-ROC}   & \textit{V-PR}  & \textit{V-ROC}    \\ \hline

\texttt{EDAD} (\textbf{ours}) & \cell \textbf{0.817} & \cell \underline{0.841} & \cell \textbf{0.829} & \cell 0.206 & \cell \underline{0.569} & \cell 0.225 & \cell \underline{0.567}  \\
\hline
w/o Stable module    & 0.805                                & \cell \textbf{0.845} & \cell 0.810 & 0.200                                & 0.568                                & \cell \textbf{0.229} & 0.557                                 \\
w/o Auxiliary module & 0.790                                & 0.826                                & 0.797                                & 0.191                                & 0.561                                & 0.212                                & 0.541                                 \\
w/o Regularization   & 0.807                                & 0.826                                & 0.806                                & \cell \textbf{0.210} & 0.558                                & 0.222                                & \cell 0.566  \\ \hline
$\circ$ \texttt{NWJ}            & 0.804                                & \cell 0.841 & \cell 0.810 & 0.205                                & 0.562                                & 0.214                                & 0.555                                 \\
$\circ$ \texttt{JSD}            & \cell 0.814 & \cell \textbf{0.845} & 0.808                                & \cell \underline{0.208} & 0.562                                & \cell 0.227 & 0.553                                 \\
$\circ$ \texttt{MINE}           & \cell \underline{0.816} & 0.828                                & \cell \underline{0.813} & 0.198                                & 0.559                                & 0.217                                & \cell \textbf{0.570}  \\ \hline
$\circ$ Bilinear       & 0.808                                & 0.836                                & 0.808                                & 0.205                                & \cell \textbf{0.573} & 0.215                                & 0.561                                 \\
$\circ$ Concatenated   & 0.813                                & 0.838                                & 0.805                                & 0.199                                & \cell \underline{0.569} & 0.215                                & 0.565 \\                               
\bottomrule
\end{tabular}
% \vspace{-20pt}
\end{table}

\subsubsection{Overall results}
\label{subsubsec:over_results}
We report on the performance of the proposed \framework and the baselines on all datasets in terms of all metrics---see Table~\ref{table:main_result}. We also report average results (see \textbf{AVERAGE}). The top 3 best results for each metric are highlighted in blue. We observe that the proposed framework achieves the top 3 highest accuracies on most datasets. According to the average results, \framework achieves the highest \textit{P}, \textit{R}, \textit{F1}, \textit{V-PR}, and \textit{V-ROC}, and it achieves the top 3 highest \textit{A-PR} and \textit{A-ROC}. This indicates the strong performance of \framework as well as significant improvements of \framework over the baselines.  

To justify whether the accuracy improvements of the proposed methods \framework over the baselines are statistically significant, we conduct $t$-tests to test the significance of the proposed methods against baselines. We consider a null hypothesis $H_{0}$ that the mean of the anomaly scores of our methods is similar to the mean of the anomaly scores of baselines, and an alternative hypothesis $H_{1}$ that the mean of the anomaly scores of our methods is different from the mean of the anomaly scores of baselines. After performing the $t$-test, we get a $p$-value, which is smaller than 0.001. This shows strong evidence to reject the null hypothesis $H_{0}$, which in turn suggests that our models have statistically outperformed baselines. 

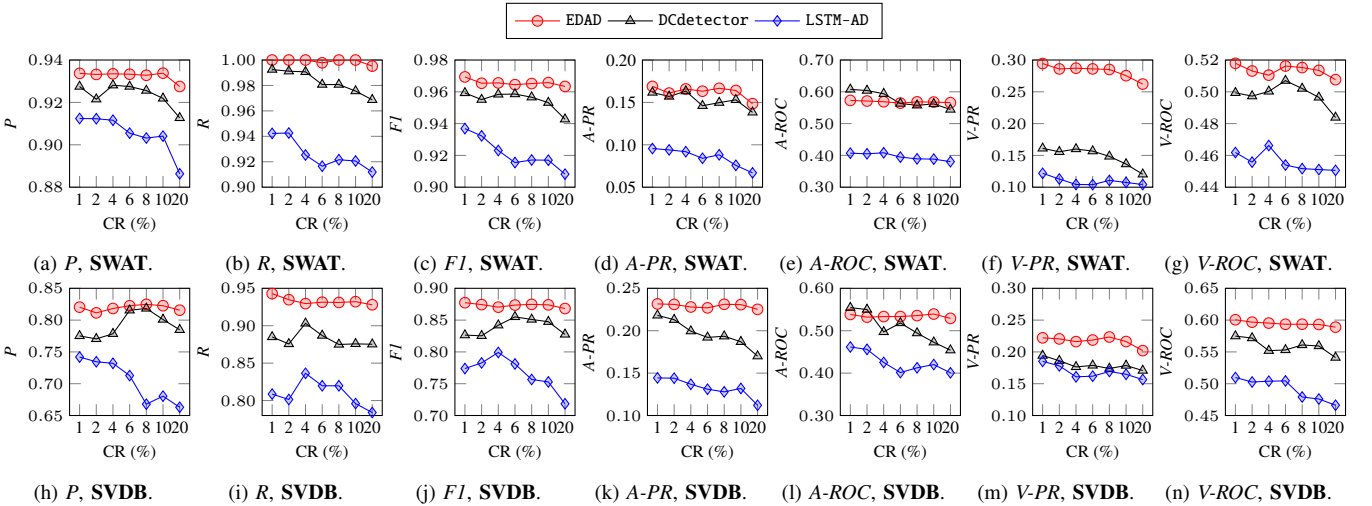
\begin{figure*}[!t]
% \vspace{-1.0em}
% \footnotesize
\fontsize{6.5pt}{6.5pt}\selectfont
\centering
\begin{subfigure}{0.138\linewidth}
    \begin{tikzpicture}
    \begin{axis}[
            xlabel=CR (\%),
            ylabel=\textit{P},
            ymax = 0.94,
            ymin = 0.88,
            xtick = {1,2,3,4,5,6,7},
            xticklabels = {1,2,4,6,8,10,20},
            yticklabel=\pgfkeys{/pgf/number format/.cd,fixed,precision=2,zerofill}\pgfmathprintnumber{\tick},
            width=1.27\linewidth,
            height=0.45*\axisdefaultheight,
            % legend pos= outer north east
            legend style={at={(5.2,1.45)},anchor=north,legend columns=-1}]
            \addplot[red,fill=none,mark=*, fill opacity=0.2] table[x=Ratio, y=EDAD-P] {data/SWAT_FULL.txt};
                \addlegendentry{\texttt{EDAD}}
            \addplot[black,fill=none,mark=triangle*, fill opacity=0.2] table[x=Ratio, y=Dcdetector-P] {data/SWAT_FULL.txt};
                \addlegendentry{\texttt{DCdetector}}
            \addplot[blue,fill=none,mark=diamond*, fill opacity=0.2] table[x=Ratio, y=LSTM-AD-P] {data/SWAT_FULL.txt};
                \addlegendentry{\texttt{LSTM-AD}} 
            % \addplot[mark=none, dashed, black, domain=2:10, samples=2] {0.92};
            % \addplot[mark=none, dashed, black, domain=2:10, samples=2] {0.915};
        \end{axis}    
    \end{tikzpicture}
    \caption{\textit{P}, \textbf{SWAT}.}
    \label{subfig:cr_p_swat}
\end{subfigure}
\begin{subfigure}{0.138\linewidth}
    \begin{tikzpicture}
        \begin{axis}[
            xlabel=CR (\%),
            ylabel=\textit{R},
            ymax = 1.00,
            ymin = 0.90,
            xtick = {1,2,3,4,5,6,7},
            xticklabels = {1,2,4,6,8,10,20},
            yticklabel=\pgfkeys{/pgf/number format/.cd,fixed,precision=2,zerofill}\pgfmathprintnumber{\tick},
            width=1.27\linewidth,
            height=0.45*\axisdefaultheight,
            % legend pos= outer north east
            legend style={at={(4.32,1.45)},anchor=north,legend columns=-1}]
            \addplot[red,fill=none,mark=*, fill opacity=0.2] table[x=Ratio, y=EDAD-R] {data/SWAT_FULL.txt};
            % \addplot[blue,fill=none,mark=+, fill opacity=0.2] table[x=Ratio, y=AnomalyTrans] {data/swat_R.txt};
            \addplot[black,fill=none,mark=triangle*, fill opacity=0.2] table[x=Ratio, y=Dcdetector-R] {data/SWAT_FULL.txt};
            \addplot[blue,fill=none,mark=diamond*, fill opacity=0.2] table[x=Ratio, y=LSTM-AD-R] {data/SWAT_FULL.txt};
            % \addplot[mark=none, dashed, black, domain=2:10, samples=2] {0.95};
            % \addplot[mark=none, dashed, black, domain=2:10, samples=2] {0.96};
        \end{axis}
    \end{tikzpicture}
    \caption{\textit{R}, \textbf{SWAT}.}
    \label{subfig:cr_r_swat}
\end{subfigure}
\begin{subfigure}{0.138\linewidth}
    \begin{tikzpicture}
        \begin{axis}[
            xlabel=CR (\%),
            ylabel=\textit{F1},
            ymax = 0.98,
            ymin = 0.90,
            xtick = {1,2,3,4,5,6,7},
            xticklabels = {1,2,4,6,8,10,20},
            yticklabel=\pgfkeys{/pgf/number format/.cd,fixed,precision=2,zerofill}\pgfmathprintnumber{\tick},
            width=1.27\linewidth,
            height=0.45*\axisdefaultheight,
            % legend pos= outer north east
            legend style={at={(4.32,1.45)},anchor=north,legend columns=-1}]
            \addplot[red,fill=none,mark=*, fill opacity=0.2] table[x=Ratio, y=EDAD-F1] {data/SWAT_FULL.txt};
            % \addplot[blue,fill=none,mark=+, fill opacity=0.2] table[x=Ratio, y=AnomalyTrans] {data/swat_F1.txt};
            \addplot[black,fill=none,mark=triangle*, fill opacity=0.2] table[x=Ratio, y=Dcdetector-F1] {data/SWAT_FULL.txt};
            \addplot[blue,fill=none,mark=diamond*, fill opacity=0.2] table[x=Ratio, y=LSTM-AD-F1] {data/SWAT_FULL.txt};
            % \addplot[mark=none, dashed, black, domain=2:10, samples=2] {0.94};
            % \addplot[mark=none, dashed, black, domain=2:10, samples=2] {0.95};
        \end{axis}
    \end{tikzpicture}
    \caption{\textit{F1}, \textbf{SWAT}.}
    \label{subfig:cr_f1_swat}
\end{subfigure}
\begin{subfigure}{0.138\linewidth}
    \begin{tikzpicture}
        \begin{axis}[
            xlabel=CR (\%),
            ylabel=\textit{A-PR},
            ymax = 0.20,
            ymin = 0.05,
            xtick = {1,2,3,4,5,6,7},
            xticklabels = {1,2,4,6,8,10,20},
            % xtick = {0.1,0.3,0.5,0.7,0.9},
            % yticklabel=\pgfkeys{/pgf/number format/.cd,fixed,precision=2,zerofill}\pgfmathprintnumber{\tick},
            ytick = {0.05, 0.10, 0.15, 0.20},
            yticklabels = {0.05, 0.10, 0.15, 0.20},
            width=1.27\linewidth,
            height=0.45*\axisdefaultheight,
            % legend pos= outer north east
            legend style={at={(4.32,1.45)},anchor=north,legend columns=-1}]
            \addplot[red,fill=none,mark=*, fill opacity=0.2] table[x=Ratio, y=EDAD-APR] {data/SWAT_FULL.txt};
            % \addplot[blue,fill=none,mark=+, fill opacity=0.2] table[x=Ratio, y=AnomalyTrans] {data/swat_F1.txt};
            \addplot[black,fill=none,mark=triangle*, fill opacity=0.2] table[x=Ratio, y=Dcdetector-APR] {data/SWAT_FULL.txt};
            \addplot[blue,fill=none,mark=diamond*, fill opacity=0.2] table[x=Ratio, y=LSTM-AD-APR] {data/SWAT_FULL.txt};
            % \addplot[mark=none, dashed, black, domain=2:10, samples=2] {0.94};
            % \addplot[mark=none, dashed, black, domain=2:10, samples=2] {0.95};
        \end{axis}
    \end{tikzpicture}
    \caption{\textit{A-PR}, \textbf{SWAT}.}
    \label{subfig:cr_apr_swat}
\end{subfigure} 
\begin{subfigure}{0.138\linewidth}
    \begin{tikzpicture}
        \begin{axis}[
            xlabel=CR (\%),
            ylabel=\textit{A-ROC},
            ymax = 0.7,
            ymin = 0.3,
            xtick = {1,2,3,4,5,6,7},
            xticklabels = {1,2,4,6,8,10,20},
            ytick = {0.30, 0.40, 0.50, 0.60, 0.70},
            yticklabels = {0.30, 0.40, 0.50, 0.60, 0.70},
            yticklabel=\pgfkeys{/pgf/number format/.cd,fixed,precision=2,zerofill}\pgfmathprintnumber{\tick},
            width=1.27\linewidth,
            height=0.45*\axisdefaultheight,
            % legend pos= outer north east
            legend style={at={(4.32,1.45)},anchor=north,legend columns=-1}]
            \addplot[red,fill=none,mark=*, fill opacity=0.2] table[x=Ratio, y=EDAD-AROC] {data/SWAT_FULL.txt};
            % \addplot[blue,fill=none,mark=+, fill opacity=0.2] table[x=Ratio, y=AnomalyTrans] {data/swat_F1.txt};
            \addplot[black,fill=none,mark=triangle*, fill opacity=0.2] table[x=Ratio, y=Dcdetector-AROC] {data/SWAT_FULL.txt};
            \addplot[blue,fill=none,mark=diamond*, fill opacity=0.2] table[x=Ratio, y=LSTM-AD-AROC] {data/SWAT_FULL.txt};
        \end{axis}
    \end{tikzpicture}
    \caption{\textit{A-ROC}, \textbf{SWAT}.}
    \label{subfig:cr_aroc_swat}
\end{subfigure}
\begin{subfigure}{0.138\linewidth}
    \begin{tikzpicture}
        \begin{axis}[
            xlabel=CR (\%),
            ylabel=\textit{V-PR},
            ymax = 0.30,
            ymin = 0.10,
            xtick = {1,2,3,4,5,6,7},
            xticklabels = {1,2,4,6,8,10,20},
            yticklabel=\pgfkeys{/pgf/number format/.cd,fixed,precision=2,zerofill}\pgfmathprintnumber{\tick},
            ytick = {0.10, 0.15, 0.20, 0.25, 0.30},
            width=1.27\linewidth,
            height=0.45*\axisdefaultheight,
            % legend pos= outer north east
            legend style={at={(4.32,1.45)},anchor=north,legend columns=-1}]
            \addplot[red,fill=none,mark=*, fill opacity=0.2] table[x=Ratio, y=EDAD-VPR] {data/SWAT_FULL.txt};
            % \addplot[blue,fill=none,mark=+, fill opacity=0.2] table[x=Ratio, y=AnomalyTrans] {data/swat_F1.txt};
            \addplot[black,fill=none,mark=triangle*, fill opacity=0.2] table[x=Ratio, y=Dcdetector-VPR] {data/SWAT_FULL.txt};
            \addplot[blue,fill=none,mark=diamond*, fill opacity=0.2] table[x=Ratio, y=LSTM-AD-VPR] {data/SWAT_FULL.txt};
        \end{axis}
    \end{tikzpicture}
    \caption{\textit{V-PR}, \textbf{SWAT}.}
    \label{subfig:cr_vpr_swat}
\end{subfigure}
\begin{subfigure}{0.138\linewidth}
    \begin{tikzpicture}
        \begin{axis}[
            xlabel=CR (\%),
            ylabel=\textit{V-ROC},
            ymax = 0.52,
            ymin = 0.44,
            xtick = {1,2,3,4,5,6,7},
            xticklabels = {1,2,4,6,8,10,20},
            yticklabel=\pgfkeys{/pgf/number format/.cd,fixed,precision=2,zerofill}\pgfmathprintnumber{\tick},
            width=1.27\linewidth,
            height=0.45*\axisdefaultheight,
            % legend pos= outer north east
            legend style={at={(4.32,1.45)},anchor=north,legend columns=-1}]
            \addplot[red,fill=none,mark=*, fill opacity=0.2] table[x=Ratio, y=EDAD-VROC] {data/SWAT_FULL.txt};
            % \addplot[blue,fill=none,mark=+, fill opacity=0.2] table[x=Ratio, y=AnomalyTrans] {data/swat_F1.txt};
            \addplot[black,fill=none,mark=triangle*, fill opacity=0.2] table[x=Ratio, y=Dcdetector-VROC] {data/SWAT_FULL.txt};
            \addplot[blue,fill=none,mark=diamond*, fill opacity=0.2] table[x=Ratio, y=LSTM-AD-VROC] {data/SWAT_FULL.txt};
            % \addplot[mark=none, dashed, black, domain=2:10, samples=2] {0.48};
            % \addplot[mark=none, dashed, black, domain=2:10, samples=2] {0.49};
        \end{axis}
    \end{tikzpicture}
    \caption{\textit{V-ROC}, \textbf{SWAT}.}
    \label{subfig:cr_vroc_swat}
\end{subfigure}

\begin{subfigure}{0.138\linewidth}
    \begin{tikzpicture}
        \begin{axis}[
            xlabel=CR (\%),
            ylabel=\textit{P},
            ymax = 0.85,
            ymin = 0.65,
            xtick = {1,2,3,4,5,6,7},
            xticklabels = {1,2,4,6,8,10,20},
            yticklabel=\pgfkeys{/pgf/number format/.cd,fixed,precision=2,zerofill}\pgfmathprintnumber{\tick},
            width=1.27\linewidth,
            height=0.45*\axisdefaultheight,
            % legend pos= outer north east
            legend style={at={(4.32,1.45)},anchor=north,legend columns=-1}]
            \addplot[red,fill=none,mark=*, fill opacity=0.2] table[x=Ratio, y=EDAD-P] {data/SVDB_FULL.txt};
            \addplot[black,fill=none,mark=triangle*, fill opacity=0.2] table[x=Ratio, y=Dcdetector-P] {data/SVDB_FULL.txt};
            \addplot[blue,fill=none,mark=diamond*, fill opacity=0.2] table[x=Ratio, y=LSTM-AD-P] {data/SVDB_FULL.txt};
            % \addplot[mark=none, dashed, black, domain=2:10, samples=2] {0.74};
            % \addplot[mark=none, dashed, black, domain=2:10, samples=2] {0.76};
        \end{axis}
    \end{tikzpicture}
    \caption{\textit{P}, \textbf{SVDB}.}
    \label{subfig:cr_p_svdb}
\end{subfigure}
\begin{subfigure}{0.138\linewidth}
    \begin{tikzpicture}
        \begin{axis}[
            xlabel=CR (\%),
            ylabel=\textit{R},
            ymax = 0.95,
            ymin = 0.78,
            xtick = {1,2,3,4,5,6,7},
            xticklabels = {1,2,4,6,8,10,20},
            yticklabel=\pgfkeys{/pgf/number format/.cd,fixed,precision=2,zerofill}\pgfmathprintnumber{\tick},
            width=1.27\linewidth,
            height=0.45*\axisdefaultheight,
            % legend pos= outer north east
            legend style={at={(4.32,1.45)},anchor=north,legend columns=-1}]
            \addplot[red,fill=none,mark=*, fill opacity=0.2] table[x=Ratio, y=EDAD-R] {data/SVDB_FULL.txt};
            \addplot[black,fill=none,mark=triangle*, fill opacity=0.2] table[x=Ratio, y=Dcdetector-R] {data/SVDB_FULL.txt};
            \addplot[blue,fill=none,mark=diamond*, fill opacity=0.2] table[x=Ratio, y=LSTM-AD-R] {data/SVDB_FULL.txt};
            % \addplot[mark=none, dashed, black, domain=2:10, samples=2] {0.84};
            % \addplot[mark=none, dashed, black, domain=2:10, samples=2] {0.86};
        \end{axis}
    \end{tikzpicture}
    \caption{\textit{R}, \textbf{SVDB}.}
    \label{subfig:cr_r_svdb}
\end{subfigure}
\begin{subfigure}{0.138\linewidth}
    \begin{tikzpicture}
        \begin{axis}[
            xlabel=CR (\%),
            ylabel=\textit{F1},
            ymax = 0.90,
            ymin = 0.70,
            xtick = {1,2,3,4,5,6,7},
            xticklabels = {1,2,4,6,8,10,20},
            yticklabel=\pgfkeys{/pgf/number format/.cd,fixed,precision=2,zerofill}\pgfmathprintnumber{\tick},
            width=1.27\linewidth,
            height=0.45*\axisdefaultheight,
            % legend pos= outer north east
            legend style={at={(4.32,1.45)},anchor=north,legend columns=-1}]
            \addplot[red,fill=none,mark=*, fill opacity=0.2] table[x=Ratio, y=EDAD-F1] {data/SVDB_FULL.txt};
            \addplot[black,fill=none,mark=triangle*, fill opacity=0.2] table[x=Ratio, y=Dcdetector-F1] {data/SVDB_FULL.txt};
            \addplot[blue,fill=none,mark=diamond*, fill opacity=0.2] table[x=Ratio, y=LSTM-AD-F1] {data/SVDB_FULL.txt};
            % \addplot[mark=none, dashed, black, domain=2:10, samples=2] {0.80};
            % \addplot[mark=none, dashed, black, domain=2:10, samples=2] {0.82};
        \end{axis}
    \end{tikzpicture}
    \caption{\textit{F1}, \textbf{SVDB}.}
    \label{subfig:cr_f1_svdb}
\end{subfigure}
\begin{subfigure}{0.138\linewidth}
    \begin{tikzpicture}
        \begin{axis}[
            xlabel=CR (\%),
            ylabel=\textit{A-PR},
            ymax = 0.25,
            ymin = 0.10,
            xtick = {1,2,3,4,5,6,7},
            xticklabels = {1,2,4,6,8,10,20},
            yticklabel=\pgfkeys{/pgf/number format/.cd,fixed,precision=2,zerofill}\pgfmathprintnumber{\tick},
            width=1.27\linewidth,
            height=0.45*\axisdefaultheight,
            % legend pos= outer north east
            legend style={at={(4.32,1.45)},anchor=north,legend columns=-1}]
            \addplot[red,fill=none,mark=*, fill opacity=0.2] table[x=Ratio, y=EDAD-APR] {data/SVDB_FULL.txt};
            \addplot[black,fill=none,mark=triangle*, fill opacity=0.2] table[x=Ratio, y=Dcdetector-APR] {data/SVDB_FULL.txt};
            \addplot[blue,fill=none,mark=diamond*, fill opacity=0.2] table[x=Ratio, y=LSTM-AD-APR] {data/SVDB_FULL.txt};
            % \addplot[mark=none, dashed, black, domain=2:10, samples=2] {0.80};
            % \addplot[mark=none, dashed, black, domain=2:10, samples=2] {0.82};
        \end{axis}
    \end{tikzpicture}
    \caption{\textit{A-PR}, \textbf{SVDB}.}
    \label{subfig:cr_apr_svdb}
\end{subfigure}
\begin{subfigure}{0.138\linewidth}
    \begin{tikzpicture}
        \begin{axis}[
            xlabel=CR (\%),
            ylabel=\textit{A-ROC},
            ymax = 0.6,
            ymin = 0.3,
            xtick = {1,2,3,4,5,6,7},
            xticklabels = {1,2,4,6,8,10,20},
            yticklabel=\pgfkeys{/pgf/number format/.cd,fixed,precision=2,zerofill}\pgfmathprintnumber{\tick},
            width=1.27\linewidth,
            height=0.45*\axisdefaultheight,
            % legend pos= outer north east
            legend style={at={(4.32,1.45)},anchor=north,legend columns=-1}]
            \addplot[red,fill=none,mark=*, fill opacity=0.2] table[x=Ratio, y=EDAD-AROC] {data/SVDB_FULL.txt};
            \addplot[black,fill=none,mark=triangle*, fill opacity=0.2] table[x=Ratio, y=Dcdetector-AROC] {data/SVDB_FULL.txt};
            \addplot[blue,fill=none,mark=diamond*, fill opacity=0.2] table[x=Ratio, y=LSTM-AD-AROC] {data/SVDB_FULL.txt};
            % \addplot[mark=none, dashed, black, domain=2:10, samples=2] {0.80};
            % \addplot[mark=none, dashed, black, domain=2:10, samples=2] {0.82};
        \end{axis}
    \end{tikzpicture}
    \caption{\textit{A-ROC}, \textbf{SVDB}.}
    \label{subfig:cr_aroc_svdb}
\end{subfigure}
\begin{subfigure}{0.138\linewidth}
    \begin{tikzpicture}
        \begin{axis}[
            xlabel=CR (\%),
            ylabel=\textit{V-PR},
            ymax = 0.3,
            ymin = 0.1,
            xtick = {1,2,3,4,5,6,7},
            xticklabels = {1,2,4,6,8,10,20},
            yticklabel=\pgfkeys{/pgf/number format/.cd,fixed,precision=2,zerofill}\pgfmathprintnumber{\tick},
            width=1.27\linewidth,
            height=0.45*\axisdefaultheight,
            % legend pos= outer north east
            legend style={at={(4.32,1.45)},anchor=north,legend columns=-1}]
            \addplot[red,fill=none,mark=*, fill opacity=0.2] table[x=Ratio, y=EDAD-VPR] {data/SVDB_FULL.txt};
            \addplot[black,fill=none,mark=triangle*, fill opacity=0.2] table[x=Ratio, y=Dcdetector-VPR] {data/SVDB_FULL.txt};
            \addplot[blue,fill=none,mark=diamond*, fill opacity=0.2] table[x=Ratio, y=LSTM-AD-VPR] {data/SVDB_FULL.txt};
            % \addplot[mark=none, dashed, black, domain=2:10, samples=2] {0.48};
            % \addplot[mark=none, dashed, black, domain=2:10, samples=2] {0.49};
        \end{axis}
    \end{tikzpicture}
    \caption{\textit{V-PR}, \textbf{SVDB}.}
    \label{subfig:cr_vpr_svdb}
\end{subfigure}
\begin{subfigure}{0.138\linewidth}
    \begin{tikzpicture}
        \begin{axis}[
            xlabel=CR (\%),
            ylabel=\textit{V-ROC},
            ymax = 0.65,
            ymin = 0.45,
            xtick = {1,2,3,4,5,6,7},
            xticklabels = {1,2,4,6,8,10,20},
            yticklabel=\pgfkeys{/pgf/number format/.cd,fixed,precision=2,zerofill}\pgfmathprintnumber{\tick},
            width=1.27\linewidth,
            height=0.45*\axisdefaultheight,
            % legend pos= outer north east
            legend style={at={(4.32,1.45)},anchor=north,legend columns=-1}]
            \addplot[red,fill=none,mark=*, fill opacity=0.2] table[x=Ratio, y=EDAD-VROC] {data/SVDB_FULL.txt};
            \addplot[black,fill=none,mark=triangle*, fill opacity=0.2] table[x=Ratio, y=Dcdetector-VROC] {data/SVDB_FULL.txt};
            \addplot[blue,fill=none,mark=diamond*, fill opacity=0.2] table[x=Ratio, y=LSTM-AD-VROC] {data/SVDB_FULL.txt};
            
        \end{axis}
    \end{tikzpicture}
    \caption{\textit{V-ROC}, \textbf{SVDB}.}
    \label{subfig:cr_vroc_svdb}
\end{subfigure}
% \vspace{-1.0em}
\caption{Effect of contamination ratio (CR).}
\label{fig:effect_contamination}
% \vspace{-10pt}
\end{figure*}

Finally, we acknowledge that it is unrealistic for a single method to be able to outperform all other methods across all datasets and metrics. In other words, there is no one-size-fits-all solution. Thus, it is unrealistic to expect our proposed method \framework to outperform all baselines in all 56 testing cases (8 datasets $\times$ 7 metrics). Among the 56 cases and when compared to 11 other methods, the proposed \framework is best in 26 cases, and second-best in 9 cases, as shown in Table~\ref{table:ranking}. The state-of-the-art method \texttt{DCdetector} is best in only 2 cases and 2nd best in 18 cases. The Compress-the-Reconstruct based method \texttt{LSTM-AD} is best in 5 cases and 2nd best in 2 cases. This clearly shows that \framework achieves superior performance.

\subsection{Ablation Study}
\label{subsec:ablation}

\subsubsection{Effect of Components}
\label{subsubsec:component}
We proceed to assess the effectiveness of each individual module in \framework. For brevity, we only report average results over the eight datasets, as shown in Table~\ref{table:main_result_v2_ablation_avg}. The results show that \framework achieves the top 3 highest accuracy when all modules are fully incorporated. If we only include a single feature module, the model with the auxiliary feature module (w/o stable feature module) can yield a better average accuracy when compared to the counterpart with the stable feature module. This suggests that the inclusion of the auxiliary feature, serving as an indicator for calculating anomaly scores, improves \framework's performance. The regularization is less important than the stable feature and the auxiliary feature modules. However, integrating the regularization into \framework can improve the performance further. In summary, the empirical findings underscore the importance of each module in \framework.

\subsubsection{Effect of Mutual Information Estimators}
\label{subsubsec:mutual_information_estimators}

We study the effect of different mutual information estimators. 
This experiment aims to characterize accurately the quality of a specific mutual information estimator, which, in turn, facilitates the accurate detection of outliers. 
Table~\ref{table:main_result_v2_ablation_avg} compares our default estimator \texttt{InfoNCE} and the state-of-the-art estimators  \texttt{NWJ}~\cite{DBLP:journals/tit/NguyenWJ10}, \texttt{MINE}~\cite{DBLP:conf/icml/BelghaziBROBHC18}, and \texttt{JSD}~\cite{DBLP:conf/icml/PooleOOAT19}. The results show that \texttt{InfoNCE} performs best, slightly ahead of \texttt{JSD}. This is because both estimators are part of the contrastive variational bounds family and treat mutual information estimation as a classification task-distinguishing joint samples from marginal ones. \texttt{InfoNCE} is a special case of \texttt{JSD} under a specific contrastive loss, and both optimize similar objectives with different bias–variance trade-offs. While their empirical performance is often comparable, \texttt{JSD} tends to be more sensitive to hyperparameters and initialization, which may affect its robustness. In addition, \texttt{NWJ} and \texttt{MINE} can also suffer from instability due to their reliance on unbounded log density ratios.

\subsubsection{Effect of Critic Functions}
While the estimators of mutual information are crucial in \framework, there is still a significant interaction between the critic function $f_\theta(\cdot)$ and the estimators. The design of the critic function determines its ability to distinguish between joint and marginal distributions. In the next experiment, we consider three commonly used critic functions, including bilinear critics~\cite{DBLP:conf/icml/PooleOOAT19}, concatenated critics~\cite{DBLP:conf/iclr/HjelmFLGBTB19}, and separable critics~\cite{DBLP:conf/nips/BachmanHB19}. Bilinear critics employ a bilinear function. Concatenated critics combine different inputs and employ a neural network to process them. Separable critics process input data in a separable manner, thus reducing computational complexity. Table~\ref{table:main_result_v2_ablation_avg} compares our default separable critic function and the two other critic functions, finding that the default separable critics perform the best. This result aligns with findings in the literature~\cite{DBLP:conf/iclr/TschannenDRGL20}.
% }

\begin{table}[t]
	% \fontsize{7.5pt}{7.5pt}\selectfont
    \footnotesize
    % \small
	\tabcolsep3pt
	\centering
	\caption{Effect of model dimensionality on training time (\textit{minutes} per \textit{epoch}).}
	\label{table:runtime}
	\begin{tabular}{c|ccc}
		\toprule
		\textbf{$d$} & \framework    & \texttt{DCdetector} & \texttt{LSTM-AD} \\
		\midrule
		128  & \textbf{1.06} & 1.07  & 1.28 \\
		256  & \textbf{1.21} & 3.04  & 1.61 \\
		512  & \textbf{1.82} & 10.94 & 2.41 \\  
		1024 & \textbf{3.64} & 41.64 & 4.55 \\
		\bottomrule
	\end{tabular}
    % \vspace{-0.5em}
\end{table}

\begin{table}[t]
	% \fontsize{7.5pt}{7.5pt}\selectfont
    \footnotesize
    % \small
	\tabcolsep3pt
	\centering
	\caption{Effect of model dimensionality on on memory cost (\textit{GB}).}
	\label{table:memory}
	\begin{tabular}{c|ccc}
		\toprule
		\textbf{$d$} & \framework   & \texttt{DCdetector} & \texttt{LSTM-AD} \\
		\midrule
		128  & 3.0 & 3.9 & \textbf{2.7} \\
		256  & \textbf{3.3} & 6.5 & 5.0 \\
		512  & \textbf{4.4} & 7.6 & 9.4 \\
		1024 & \textbf{6.4} & 10.1 & 18.4 \\
		\bottomrule
	\end{tabular}
    % \vspace{-1.5em}
\end{table}

\input{charts/lambda}
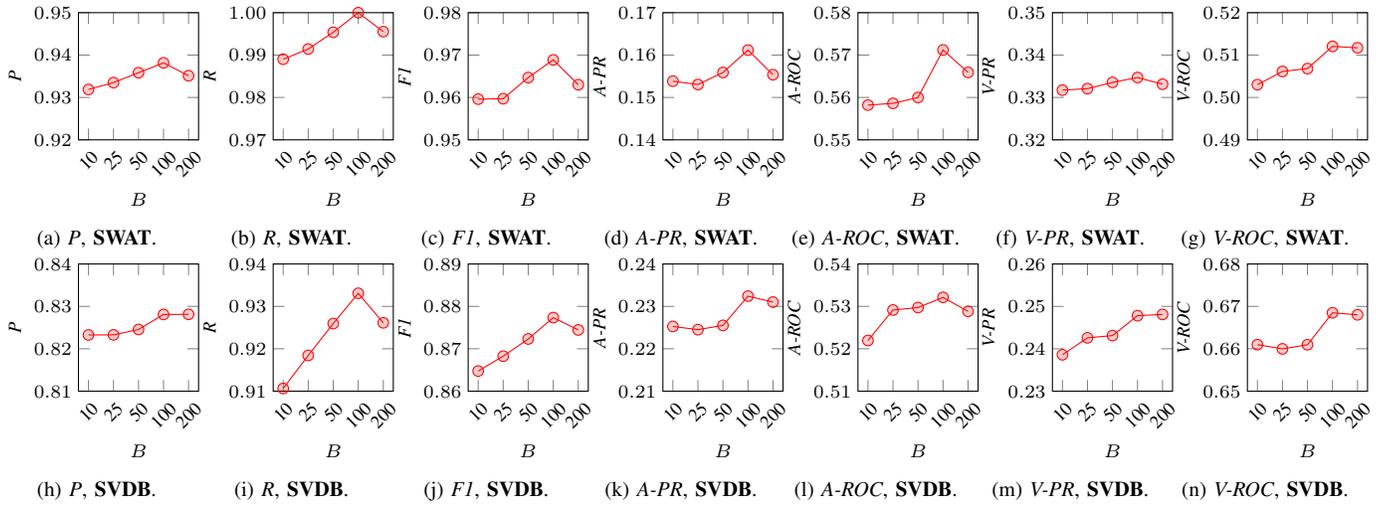
\begin{figure*}[!t]
% \vspace{-1.0em}
% \footnotesize
\fontsize{7.0pt}{7.0pt}\selectfont
\centering

\begin{subfigure}{0.138\linewidth}
    \begin{tikzpicture}
        \begin{axis}[
            xlabel=$B$,
            ylabel=\textit{P},
            ymax = 0.95,
            ymin = 0.92,
            xtick = {1, 2, 3, 4, 5},
            xticklabels={10, 25, 50, 100, 200 }, 
            xticklabel style={rotate=45},
            yticklabel=\pgfkeys{/pgf/number format/.cd,fixed,precision=2,zerofill}\pgfmathprintnumber{\tick},
            ytick = {0.92, 0.93, 0.94, 0.95},
            width=1.27\linewidth,
            height=0.45*\axisdefaultheight,
            % legend pos= outer north east
            legend style={at={(6.0,1.45)},anchor=north,legend columns=-1}]
            \addplot[red,fill=none,mark=*, fill opacity=0.2] table[x=Ratio, y=MIDD] {data/win_size/SWAT_P.txt};
        \end{axis}
    \end{tikzpicture}
    \caption{\textit{P}, \textbf{SWAT}.}
    \label{subfig:window_cr_p_swat}
\end{subfigure}
\begin{subfigure}{0.138\linewidth}
    \begin{tikzpicture}
        \begin{axis}[
            xlabel=$B$,
            ylabel=\textit{R},
            ymax = 1,
            ymin = 0.97,
            xtick = {1, 2, 3, 4, 5},
            xticklabels={10, 25, 50, 100, 200 }, 
            xticklabel style={rotate=45},
            yticklabel=\pgfkeys{/pgf/number format/.cd,fixed,precision=2,zerofill}\pgfmathprintnumber{\tick},
            ytick = {0.97, 0.98, 0.99, 1.00},
            width=1.27\linewidth,
            height=0.45*\axisdefaultheight,
            % legend pos= outer north east
            legend style={at={(6.0,1.45)},anchor=north,legend columns=-1}]
            \addplot[red,fill=none,mark=*, fill opacity=0.2] table[x=Ratio, y=MIDD] {data/win_size/SWAT_R.txt};
            % \addplot[blue,fill=none,mark=+, fill opacity=0.2] table[x=Ratio, y=AnomalyTrans] {data/win size/SWAT_R.txt};
            % \addplot[black,fill=none,mark=triangle*, fill opacity=0.2] table[x=Ratio, y=DCdetector] {data/win_size/SWAT_R.txt};
            % \addplot[blue,fill=none,mark=diamond*, fill opacity=0.2] table[x=Ratio, y=TimesNet] {data/win_size/SWAT_R.txt};
        \end{axis}
    \end{tikzpicture}
    \caption{\textit{R}, \textbf{SWAT}.}
    \label{subfig:window_cr_r_swat}
\end{subfigure}
\begin{subfigure}{0.138\linewidth}
    \begin{tikzpicture}
        \begin{axis}[
            xlabel=$B$,
            ylabel=\textit{F1},
            ymax = 0.98,
            ymin = 0.95,
            xtick = {1, 2, 3, 4, 5},
            xticklabels={10, 25, 50, 100, 200}, 
            xticklabel style={rotate=45},
            yticklabel=\pgfkeys{/pgf/number format/.cd,fixed,precision=2,zerofill}\pgfmathprintnumber{\tick},
            ytick = {0.95, 0.96, 0.97, 0.98},
            width=1.27\linewidth,
            height=0.45*\axisdefaultheight,
            % legend pos= outer north east
            legend style={at={(6.0,1.45)},anchor=north,legend columns=-1}]
            \addplot[red,fill=none,mark=*, fill opacity=0.2] table[x=Ratio, y=MIDD] {data/win_size/SWAT_F1.txt};
            % \addplot[blue,fill=none,mark=+, fill opacity=0.2] table[x=Ratio, y=AnomalyTrans] {data/win size/SWAT_F1.txt};
            % \addplot[black,fill=none,mark=triangle*, fill opacity=0.2] table[x=Ratio, y=DCdetector] {data/win size/SWAT_F1.txt};
            % \addplot[blue,fill=none,mark=diamond*, fill opacity=0.2] table[x=Ratio, y=TimesNet] {data/win size/SWAT_F1.txt};
        \end{axis}
    \end{tikzpicture}
    \caption{\textit{F1}, \textbf{SWAT}.}
    \label{subfig:window_cr_f1_swat}
\end{subfigure}
\begin{subfigure}{0.138\linewidth}
    \begin{tikzpicture}
        \begin{axis}[
            xlabel=$B$,
            ylabel=\textit{A-PR},
            ymax = 0.17,
            ymin = 0.14,
            xtick = {1, 2, 3, 4, 5},
            xticklabels={10, 25, 50, 100, 200 }, 
            xticklabel style={rotate=45},
            yticklabel=\pgfkeys{/pgf/number format/.cd,fixed,precision=2,zerofill}\pgfmathprintnumber{\tick},
            ytick = {0.14, 0.15, 0.16, 0.17},
            width=1.27\linewidth,
            height=0.45*\axisdefaultheight,
            % legend pos= outer north east
            legend style={at={(6.0,1.45)},anchor=north,legend columns=-1}]
            \addplot[red,fill=none,mark=*, fill opacity=0.2] table[x=Ratio, y=MIDD] {data/win_size/SWAT_APR.txt};
            % \addplot[blue,fill=none,mark=+, fill opacity=0.2] table[x=Ratio, y=AnomalyTrans] {data/win size/SWAT_F1.txt};
            % \addplot[black,fill=none,mark=triangle*, fill opacity=0.2] table[x=Ratio, y=DCdetector] {data/win size/SWAT_APR.txt};
            % \addplot[blue,fill=none,mark=diamond*, fill opacity=0.2] table[x=Ratio, y=TimesNet] {data/win size/SWAT_APR.txt};
        \end{axis}
    \end{tikzpicture}
    \caption{\textit{A-PR}, \textbf{SWAT}.}
    \label{subfig:window_cr_apr_swat}
\end{subfigure} 
\begin{subfigure}{0.138\linewidth}
    \begin{tikzpicture}
        \begin{axis}[
            xlabel=$B$,
            ylabel=\textit{A-ROC},
            ymax = 0.58,
            ymin = 0.55,
            xtick = {1, 2, 3, 4, 5},
            xticklabels={10, 25, 50, 100, 200 }, 
            xticklabel style={rotate=45},
            yticklabel=\pgfkeys{/pgf/number format/.cd,fixed,precision=2,zerofill}\pgfmathprintnumber{\tick},
            ytick = {0.55, 0.56, 0.57, 0.58},
            width=1.27\linewidth,
            height=0.45*\axisdefaultheight,
            % legend pos= outer north east
            legend style={at={(6.0,1.45)},anchor=north,legend columns=-1}]
            \addplot[red,fill=none,mark=*, fill opacity=0.2] table[x=Ratio, y=MIDD] {data/win_size/SWAT_AROC.txt};
            % \addplot[blue,fill=none,mark=+, fill opacity=0.2] table[x=Ratio, y=AnomalyTrans] {data/win size/SWAT_F1.txt};
            % \addplot[black,fill=none,mark=triangle*, fill opacity=0.2] table[x=Ratio, y=DCdetector] {data/win size/SWAT_AROC.txt};
            % \addplot[blue,fill=none,mark=diamond*, fill opacity=0.2] table[x=Ratio, y=TimesNet] {data/win size/SWAT_AROC.txt};
        \end{axis}
    \end{tikzpicture}
    \caption{\textit{A-ROC}, \textbf{SWAT}.}
    \label{subfig:window_cr_aroc_swat}
\end{subfigure}
\begin{subfigure}{0.138\linewidth}
    \begin{tikzpicture}
        \begin{axis}[
            xlabel=$B$,
            ylabel=\textit{V-PR},
            ymax = 0.35,
            ymin = 0.32,
            xtick = {1, 2, 3, 4, 5},
            xticklabels={10, 25, 50, 100, 200 }, 
            xticklabel style={rotate=45},
            yticklabel=\pgfkeys{/pgf/number format/.cd,fixed,precision=2,zerofill}\pgfmathprintnumber{\tick},
            ytick = {0.32, 0.33, 0.34, 0.35},
            width=1.27\linewidth,
            height=0.45*\axisdefaultheight,
            % legend pos= outer north east
            legend style={at={(6.0,1.45)},anchor=north,legend columns=-1}]
            \addplot[red,fill=none,mark=*, fill opacity=0.2] table[x=Ratio, y=MIDD] {data/win_size/SWAT_VPR.txt};
            % \addplot[blue,fill=none,mark=+, fill opacity=0.2] table[x=Ratio, y=AnomalyTrans] {data/win size/SWAT_F1.txt};
            % \addplot[black,fill=none,mark=triangle*, fill opacity=0.2] table[x=Ratio, y=DCdetector] {data/win size/SWAT_VPR.txt};
            % \addplot[blue,fill=none,mark=diamond*, fill opacity=0.2] table[x=Ratio, y=TimesNet] {data/win size/SWAT_VPR.txt};
        \end{axis}
    \end{tikzpicture}
    \caption{\textit{V-PR}, \textbf{SWAT}.}
    \label{subfig:window_cr_vpr_swat}
\end{subfigure}
\begin{subfigure}{0.138\linewidth}
    \begin{tikzpicture}
        \begin{axis}[
            xlabel=$B$,
            ylabel=\textit{V-ROC},
            ymax = 0.52,
            ymin = 0.49,
            xtick = {1, 2, 3, 4, 5},
            xticklabels={10, 25, 50, 100, 200 }, 
            xticklabel style={rotate=45},
            yticklabel=\pgfkeys{/pgf/number format/.cd,fixed,precision=2,zerofill}\pgfmathprintnumber{\tick},
            ytick = {0.49, 0.50, 0.51, 0.52},
            width=1.27\linewidth,
            height=0.45*\axisdefaultheight,
            % legend pos= outer north east
            legend style={at={(6.0,1.45)},anchor=north,legend columns=-1}]
            \addplot[red,fill=none,mark=*, fill opacity=0.2] table[x=Ratio, y=MIDD] {data/win_size/SWAT_VROC.txt};
            % \addplot[blue,fill=none,mark=+, fill opacity=0.2] table[x=Ratio, y=AnomalyTrans] {data/win size/SWAT_F1.txt};
            % \addplot[black,fill=none,mark=triangle*, fill opacity=0.2] table[x=Ratio, y=DCdetector] {data/win size/SWAT_VROC.txt};
            % \addplot[blue,fill=none,mark=diamond*, fill opacity=0.2] table[x=Ratio, y=TimesNet] {data/win size/SWAT_VROC.txt};
        \end{axis}
    \end{tikzpicture}
    \caption{\textit{V-ROC}, \textbf{SWAT}.}
    \label{subfig:window_cr_vroc_swat}
\end{subfigure}

\begin{subfigure}{0.138\linewidth}
    \begin{tikzpicture}
        \begin{axis}[
            xlabel=$B$,
            ylabel=\textit{P},
            ymax = 0.84,
            ymin = 0.81,
            xtick = {1, 2, 3, 4, 5},
            xticklabels={10, 25, 50, 100, 200}, 
            xticklabel style={rotate=45},
            yticklabel=\pgfkeys{/pgf/number format/.cd,fixed,precision=2,zerofill}\pgfmathprintnumber{\tick},
            ytick = {0.81, 0.82, 0.83, 0.84},
            width=1.27\linewidth,
            height=0.45*\axisdefaultheight,
            % legend pos= outer north east
            legend style={at={(6.0,1.45)},anchor=north,legend columns=-1}]
            \addplot[red,fill=none,mark=*, fill opacity=0.2] table[x=Ratio, y=MIDD] {data/win_size/SVDB_P.txt};
            % \addplot[blue,fill=none,mark=+, fill opacity=0.2] table[x=Ratio, y=AnomalyTrans] {data/win size/SWAT_P.txt};
            % \addplot[black,fill=none,mark=triangle*, fill opacity=0.2] table[x=Ratio, y=DCdetector] {data/win_size/SVDB_P.txt};
            % \addplot[blue,fill=none,mark=diamond*, fill opacity=0.2] table[x=Ratio, y=TimesNet] {data/win_size/SVDB_P.txt};
        \end{axis}
    \end{tikzpicture}
    \caption{\textit{P}, \textbf{SVDB}.}
    \label{subfig:window_cr_p_svdb}
\end{subfigure}
\begin{subfigure}{0.138\linewidth}
    \begin{tikzpicture}
        \begin{axis}[
            xlabel=$B$,
            ylabel=\textit{R},
            ymax = 0.94,
            ymin = 0.91,
            xtick = {1, 2, 3, 4, 5},
            xticklabels={10, 25, 50, 100, 200 }, 
            xticklabel style={rotate=45},
            yticklabel=\pgfkeys{/pgf/number format/.cd,fixed,precision=2,zerofill}\pgfmathprintnumber{\tick},
            ytick = {0.91, 0.92, 0.93, 0.94},
            width=1.27\linewidth,
            height=0.45*\axisdefaultheight,
            % legend pos= outer north east
            legend style={at={(6.0,1.45)},anchor=north,legend columns=-1}]
            \addplot[red,fill=none,mark=*, fill opacity=0.2] table[x=Ratio, y=MIDD] {data/win_size/SVDB_R.txt};
            % \addplot[blue,fill=none,mark=+, fill opacity=0.2] table[x=Ratio, y=AnomalyTrans] {data/win size/SWAT_R.txt};
            % \addplot[black,fill=none,mark=triangle*, fill opacity=0.2] table[x=Ratio, y=DCdetector] {data/win_size/SVDB_R.txt};   
            % \addplot[blue,fill=none,mark=diamond*, fill opacity=0.2] table[x=Ratio, y=TimesNet] {data/win_size/SVDB_R.txt};
        \end{axis}
    \end{tikzpicture}
    \caption{\textit{R}, \textbf{SVDB}.}
    \label{subfig:window_cr_r_svdb}
\end{subfigure}
\begin{subfigure}{0.138\linewidth}
    \begin{tikzpicture}
        \begin{axis}[
            xlabel=$B$,
            ylabel=\textit{F1},
            ymax = 0.89,
            ymin = 0.86,
            xtick = {1, 2, 3, 4, 5},
            xticklabels={10, 25, 50, 100, 200 }, 
            xticklabel style={rotate=45},
            yticklabel=\pgfkeys{/pgf/number format/.cd,fixed,precision=2,zerofill}\pgfmathprintnumber{\tick},
            ytick = {0.86, 0.87, 0.88, 0.89},
            width=1.27\linewidth,
            height=0.45*\axisdefaultheight,
            % legend pos= outer north east
            legend style={at={(6.0,1.45)},anchor=north,legend columns=-1}]
            \addplot[red,fill=none,mark=*, fill opacity=0.2] table[x=Ratio, y=MIDD] {data/win_size/SVDB_F1.txt};
            % \addplot[blue,fill=none,mark=+, fill opacity=0.2] table[x=Ratio, y=AnomalyTrans] {data/win size/SWAT_F1.txt};
            % \addplot[black,fill=none,mark=triangle*, fill opacity=0.2] table[x=Ratio, y=DCdetector] {data/win_size/SVDB_F1.txt};
            % \addplot[blue,fill=none,mark=diamond*, fill opacity=0.2] table[x=Ratio, y=TimesNet] {data/win_size/SVDB_F1.txt};
        \end{axis}b
    \end{tikzpicture}
    \caption{\textit{F1}, \textbf{SVDB}.}
    \label{subfig:window_cr_f1_svdb}
\end{subfigure}
\begin{subfigure}{0.138\linewidth}
    \begin{tikzpicture}
        \begin{axis}[
            xlabel=$B$,
            ylabel=\textit{A-PR},
            ymax = 0.24,
            ymin = 0.21,
            xtick = {1, 2, 3, 4, 5},
            xticklabels={10, 25, 50, 100, 200 }, 
            xticklabel style={rotate=45},
            yticklabel=\pgfkeys{/pgf/number format/.cd,fixed,precision=2,zerofill}\pgfmathprintnumber{\tick},
            ytick = {0.21, 0.22, 0.23, 0.24},
            width=1.27\linewidth,
            height=0.45*\axisdefaultheight,
            % legend pos= outer north east
            legend style={at={(6.0,1.45)},anchor=north,legend columns=-1}]
            \addplot[red,fill=none,mark=*, fill opacity=0.2] table[x=Ratio, y=MIDD] {data/win_size/SVDB_APR.txt};
            % \addplot[blue,fill=none,mark=+, fill opacity=0.2] table[x=Ratio, y=AnomalyTrans] {data/win size/SWAT_F1.txt};
            % \addplot[black,fill=none,mark=triangle*, fill opacity=0.2] table[x=Ratio, y=DCdetector] {data/win_size/SVDB_APR.txt};
            % \addplot[blue,fill=none,mark=diamond*, fill opacity=0.2] table[x=Ratio, y=TimesNet] {data/win_size/SVDB_APR.txt};
        \end{axis}
    \end{tikzpicture}
    \caption{\textit{A-PR}, \textbf{SVDB}.}
    \label{subfig:window_cr_apr_svdb}
\end{subfigure}
\begin{subfigure}{0.138\linewidth}
    \begin{tikzpicture}
        \begin{axis}[
            xlabel=$B$,
            ylabel=\textit{A-ROC},
            ymax = 0.54,
            ymin = 0.51,
            xtick = {1, 2, 3, 4, 5},
            xticklabels={10, 25, 50, 100, 200 }, 
            xticklabel style={rotate=45},
            yticklabel=\pgfkeys{/pgf/number format/.cd,fixed,precision=2,zerofill}\pgfmathprintnumber{\tick},
            ytick = {0.51, 0.52, 0.53, 0.54},
            width=1.27\linewidth,
            height=0.45*\axisdefaultheight,
            % legend pos= outer north east
            legend style={at={(6.0,1.45)},anchor=north,legend columns=-1}]
            \addplot[red,fill=none,mark=*, fill opacity=0.2] table[x=Ratio, y=MIDD] {data/win_size/SVDB_AROC.txt};
            % \addplot[blue,fill=none,mark=+, fill opacity=0.2] table[x=Ratio, y=AnomalyTrans] {data/win size/SWAT_F1.txt};
            % \addplot[black,fill=none,mark=triangle*, fill opacity=0.2] table[x=Ratio, y=DCdetector] {data/win_size/SVDB_AROC.txt};
            % \addplot[blue,fill=none,mark=diamond*, fill opacity=0.2] table[x=Ratio, y=TimesNet] {data/win_size/SVDB_AROC.txt};
        \end{axis}
    \end{tikzpicture}
    \caption{\textit{A-ROC}, \textbf{SVDB}.}
    \label{subfig:window_cr_aroc_svdb}
\end{subfigure}
\begin{subfigure}{0.138\linewidth}
    \begin{tikzpicture}
        \begin{axis}[
            xlabel=$B$,
            ylabel=\textit{V-PR},
            ymax = 0.26,
            ymin = 0.23,
            xtick = {1, 2, 3, 4, 5},
            xticklabels={10, 25, 50, 100, 200 }, 
            xticklabel style={rotate=45},
            yticklabel=\pgfkeys{/pgf/number format/.cd,fixed,precision=2,zerofill}\pgfmathprintnumber{\tick},
            ytick = {0.23, 0.24, 0.25, 0.26},
            width=1.27\linewidth,
            height=0.45*\axisdefaultheight,
            % legend pos= outer north east
            legend style={at={(6.0,1.45)},anchor=north,legend columns=-1}]
            \addplot[red,fill=none,mark=*, fill opacity=0.2] table[x=Ratio, y=MIDD] {data/win_size/SVDB_VPR.txt};
            % \addplot[blue,fill=none,mark=+, fill opacity=0.2] table[x=Ratio, y=AnomalyTrans] {data/win size/SWAT_F1.txt};
            % \addplot[black,fill=none,mark=triangle*, fill opacity=0.2] table[x=Ratio, y=DCdetector] {data/win_size/SVDB_PR.txt};
            % \addplot[blue,fill=none,mark=diamond*, fill opacity=0.2] table[x=Ratio, y=TimesNet] {data/win_size/SVDB_PR.txt};
        \end{axis}
    \end{tikzpicture}
    \caption{\textit{V-PR}, \textbf{SVDB}.}
    \label{subfig:window_cr_vpr_svdb}
\end{subfigure}
\begin{subfigure}{0.138\linewidth}
    \begin{tikzpicture}
        \begin{axis}[
            xlabel=$B$,
            ylabel=\textit{V-ROC},
            ymax = 0.68,
            ymin = 0.65,
            xtick = {1, 2, 3, 4, 5},
            xticklabels={10, 25, 50, 100, 200 }, 
            xticklabel style={rotate=45},
            yticklabel=\pgfkeys{/pgf/number format/.cd,fixed,precision=2,zerofill}\pgfmathprintnumber{\tick},
            ytick = {0.65, 0.66, 0.67, 0.68},
            width=1.27\linewidth,
            height=0.45*\axisdefaultheight,
            % legend pos= outer north east
            legend style={at={(6.0,1.45)},anchor=north,legend columns=-1}]
            \addplot[red,fill=none,mark=*, fill opacity=0.2] table[x=Ratio, y=MIDD] {data/win_size/SVDB_VROC.txt};
            % \addplot[blue,fill=none,mark=+, fill opacity=0.2] table[x=Ratio, y=AnomalyTrans] {data/win size/SWAT_F1.txt};
            % \addplot[black,fill=none,mark=triangle*, fill opacity=0.2] table[x=Ratio, y=DCdetector] {data/win_size/SVDB_ROC.txt};
            % \addplot[blue,fill=none,mark=diamond*, fill opacity=0.2] table[x=Ratio, y=TimesNet] {data/win_size/SVDB_ROC.txt};
            
        \end{axis}
    \end{tikzpicture}
    \caption{\textit{V-ROC}, \textbf{SVDB}.}
    \label{subfig:window_cr_vroc_svdb}
\end{subfigure}
% \captionsetup{skip=20pt}
\caption{Effect of $B$.}
\label{fig:effect_win}
% \vspace{-10pt}
\end{figure*}

\subsubsection{Contamination Robustness}
\label{subsubsec:robustness}
We aim to evaluate the robustness of a method at different levels of contamination. To enable this experiment, we modify a proportion of the original observations and consider the modified observations as anomalies \cite{DBLP:conf/iclr/GoswamiCCMK23, DBLP:journals/corr/abert}. We vary the anomaly ratio among 1\%, 2\%, 4\%, 6\%, 8\%, 10\%, and 20\%. For brevity, we conduct experiments on two datasets: \textbf{SWAT} and \textbf{SVDB}, and we compare \framework with two methods: 1) \texttt{LSTM-AE}, which is a reconstruction-based method employing a Compress-then-Reconstruct paradigm, and 2) \texttt{DCdetector}, which is a robust anomaly detection method. We acknowledge that injected anomalies may not fully reflect the complexity of noise found in real-world contaminated data. However, they still serve as a useful proxy for evaluating the robustness of anomaly detection methods. Figure~\ref{fig:effect_contamination} shows the experimental results. We observe that \texttt{DCdetector} performs well with a competitive result due to its ability to learn robust representations by using contrastive learning. However, \texttt{DCdetector} achieves an inferior performance to \framework. This demonstrates that \texttt{DCdetector} is less robust than \framework. The results show that \framework outperforms \texttt{LSTM-AE} w.r.t. all metrics. When the contamination ratio increases, \framework maintains good performance w.r.t. all metrics. In contrast, \texttt{LSTM-AE} tends to exhibit serious drops in performance. This suggests that \framework is able to work on contaminated data with performance that is insensitive to the level of contamination.

\subsubsection{Runtime Analysis}
\label{subsubsec:runtime}
To study the deployment potential of \framework, we compare its runtime (i.e., online detection time) with two methods in previous experiments: \texttt{DCdetector} and \texttt{LSTM-AD}. First, we determine the runtime on each dataset. Then, we report the average runtime over all datasets. To achieve fair comparisons, we keep the dimensionality of the hidden states $d$ the same across the methods. We also observe that the runtime mainly comes from the offline training time. Table~\ref{table:runtime} reports the time needed (in \textit{minutes}) to finish one training \textit{epoch}. The highest results are highlighted with bold text.

The training time results show that \framework performs the fastest, whereas \texttt{DCdetector} runs much slower. This is because \texttt{DCdetector} has a dual attention component whereas \framework employs only a single attention. Further, the results show that \framework is able to train in a very short time. The online detection time of \framework is small, i.e., less than 0.1 second, making it applicable to online anomaly detection in streaming settings.

\subsubsection{Memory Analysis}
\label{subsubsec:memory}
We study the memory consumption of \framework and compare it with the memory consumption of two methods in previous experiments: \texttt{DCdetector} and \texttt{LSTM-AD}. 
First, we determine the memory consumption on each dataset. Then, we report the average memory consumption over all datasets. 
To achieve fair comparisons, we keep the dimensionality of the hidden states $d$ the same across the methods. 
Table~\ref{table:memory} shows the RAM (in \textit{GB}) used by the methods for training. The best results are highlighted with bold text.

We observe that \framework consumes the least memory in most cases except the case d = 128, and \texttt{DCdetector} consumes the most memory. 
This is because \texttt{DCdetector} has a dual attention component whereas \framework only uses single attention. 
This suggests that \framework is able to perform on low-cost off-the-shelf computers. 
This enables the use of \framework in many different resource-limited environments.

\subsubsection{Effect of $\lambda_{1}$, $\lambda_{2}$, and $\lambda_{3}$}
\label{subsubsec:lambda_effect}
We study the sensitivity of the hyperparameters $\lambda_{1}$, $\lambda_{2}$, and $\lambda_{3}$ in the objective function of \framework (see Eq. 25). 
Specifically, we vary one $\lambda$ among 0.1, 0.5, 1, 2, and 3 while keeping the other two fixed at 1 to investigate the sensitivity to the hyper-parameter. 
Figure~\ref{fig:effect_lambda} shows the results.
First, $\lambda_{3}$ controls the strength of the regularization loss. 
In most cases, as it increases, the model’s performance decreases gradually. 
This indicates that an excessively high regularization strength can hinder representation learning. 
Second, $\lambda_{1}$ and $\lambda_{2}$ control the trade-off between the two novel modules in \framework. 
They mutually learn different features in the representations of time series. 
We observe that when the weights of the two modules are approximately equal, the model achieves the best performance in most cases. 
This is evidence that the stable and auxiliary modules are equally important and indispensable components of \framework.

\subsubsection{Effect of window size $B$}
\label{subsubsec:win_size_effect}
We study the effect of window size $B$. 
More specifically, we vary $B$ among 10, 25, 50, 100, and 200 to investigate the sensitivity to $B$. 
Figure~\ref{fig:effect_win} shows the experimental results. 
We observe that when $B$ increases, the model's performance increases gradually and becomes stable with $B \ge 50$. 
In many cases, the model's performance achieves the peak with $B = 100$. 
Then, the model's performance starts to decrease with $B > 100$. 
This observation aligns with existing studies~\cite{DBLP:conf/www/XuCZLBLLZPFCWQ18,DBLP:conf/kdd/SuZNLSP19}, where they claim that deep anomaly detection methods frequently achieve the best accuracy with $B$ is set around 100.  

\begin{figure*}[!t]
    % \vspace{-0.5em}
    \centering
    \includegraphics[width=0.85\linewidth]{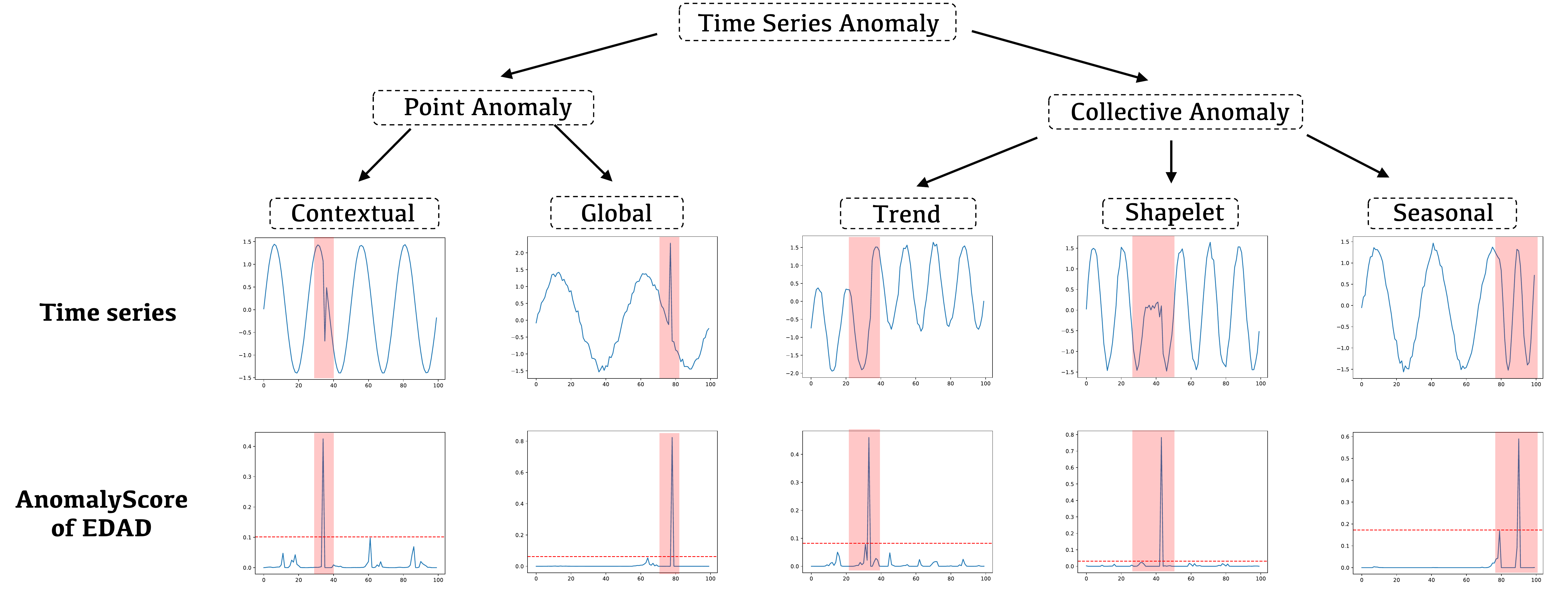}
    \caption{Visualization of case study on anomalies built upon from Lai et al.~\cite{DBLP:conf/nips/LaiZXZWH21}. \textit{Global} anomalies and \textit{contextual} anomalies are types of \textit{point} anomalies. Global anomalies refer to data points that significantly deviate from the normal pattern of the entire time series, while contextual anomalies are data points that are considered abnormal only in a specific context. \textit{Shapelet} anomalies, \textit{seasonal} anomalies, and \textit{trend} anomalies belong to the type of \textit{collective} anomaly. Shapelet anomalies refer to a subsequence within the data whose shape is inconsistent with the normal pattern of the entire time series. Seasonal anomalies refer to abnormalities in the seasonal pattern of the data. Trend anomalies are patterns that contradict the long-term trend of a time series.} 
    \label{fig:case_study}
    % \vspace{-1.5em}
\end{figure*}

\subsection{Visualization} 
\label{subsec:visualization}

\begin{figure}[t]
    \centering
    \includegraphics[width=1.0\linewidth]{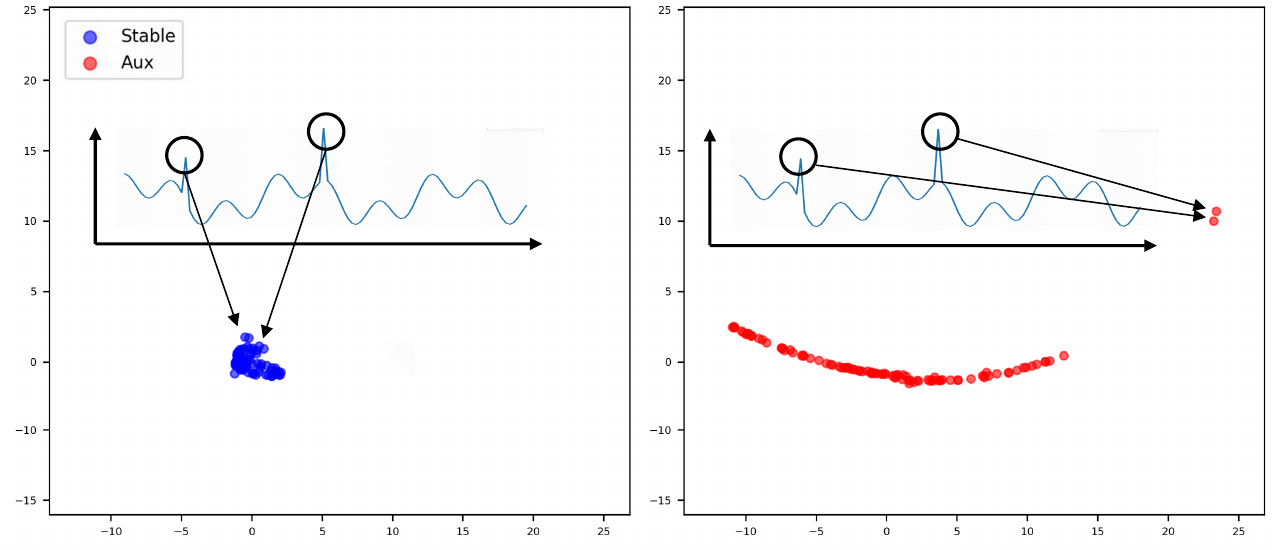}
    \caption{An example of the distribution of stable features and auxiliary features.}
    \label{fig:rf1}
    % \vspace{-1.5em}
\end{figure}

\begin{figure}[t]
    \centering
    \includegraphics[width=1.0\linewidth]{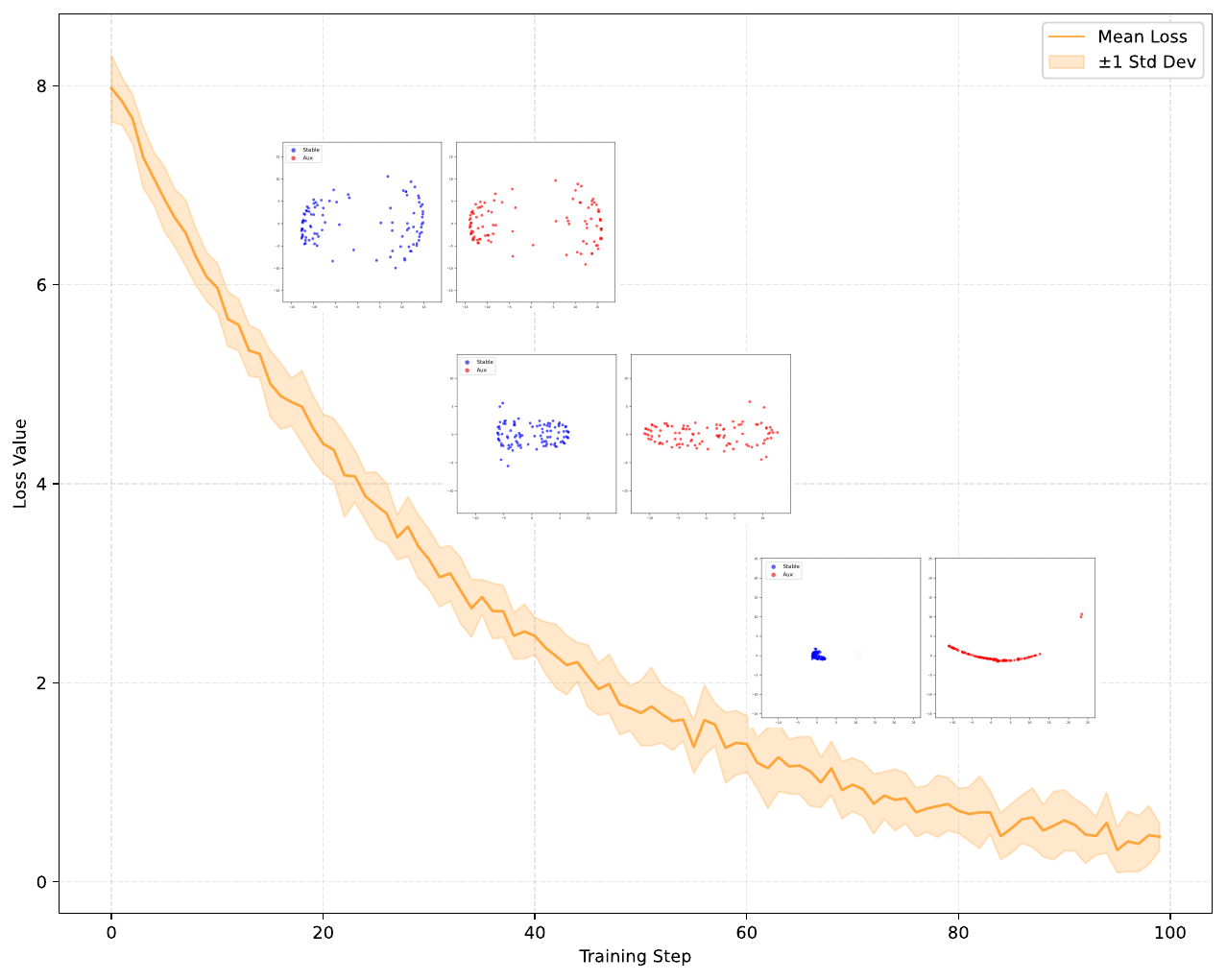}
    \caption{Training dynamics. Auxiliary features and stable features use the same axis scale.}
    \label{fig:rf2}
    % \vspace{-1.5em}
\end{figure}

In order to offer a more comprehensible and intuitive illustration of how our method excels in detecting diverse anomalies within time series data, we intentionally designed and generated various types of anomaly sequences. The primary aim was to visually showcase the model’s proficiency in identifying anomalies across different categories. Building upon the categorization of anomaly types as summarized in the work by Lai et al.~\cite{DBLP:conf/nips/LaiZXZWH21}, we subjected our method to the assessment of five specific anomaly types: \textit{global}, \textit{contextual}, \textit{shapelet}, \textit{seasonal}, and \textit{trend}. Figure~\ref{fig:case_study} visualizes the detection results. While the figure is adapted from Lai et al., it accurately reflects the characteristics observed by using our proposed method, which is why we chose to include it. The results demonstrate that our proposed framework can detect different types of anomalies. This offers evidence of the effectiveness of our approach and its capability to work on practical problems where different anomaly types occur.

Next, we empirically analyze stable and auxiliary features to further illustrate and understand their sensitivity. Figure~\ref{fig:rf1} visualizes the distribution of stable and auxiliary features for a toy dataset in two separate low-dimensional spaces using \texttt{t-SNE}~\cite{van2008visualizing}. Note that we use the same axis scale for the visualization of auxiliary features and the visualization of stable features. It is clear that the distribution of auxiliary features is more dispersed than that of stable features. Recall that we use different strategies in the proposed stable module and auxiliary module, which leads to different representations. For stable features, we assume that the normal pattern in the time series is persistent and is the normal form of the data, so the distribution of stable features is relatively concentrated. For auxiliary features, they contain noise and anomalies related to timestamps, and this randomness results in the features being relatively dispersed. Furthermore, Figure~\ref{fig:rf2} illustrates how the distributions of stable and auxiliary features evolve across training epochs, again visualized with \texttt{t-SNE}. Initially, the stable and auxiliary features exhibit similar distributions. However, as the training progresses, the distinction between these two feature types becomes increasingly pronounced. By the final stages of training, the auxiliary feature representation provides a clearer separation between normal instances and anomalies, whereas the stable feature representation fails to distinguish them.

\section{Related Work}
\label{sec:related}
\noindent\textbf{Time Series Anomaly Detection.}
Many time series anomaly detection approaches exist, including traditional statistical methods, classical machine learning algorithms~\cite{DBLP:conf/edbt/Senin0WOGBCF15,DBLP:conf/sigmod/BreunigKNS00}, and modern deep learning methods. 
Traditional statistical methods detect anomalies by applying an auto-regression mechanism~\cite{Chatfield_1978,DBLP:conf/sera/DuMLLSL18,mahimkar2011rapid}. These methods are easy to implement and deploy. However, their accuracy is relatively low. 
Classical machine algorithms can be categorized into similarity-based and density-based methods. In similarity-based methods, time series subsequences are compared. The most different subsequences are likely to be anomalies. Senin et al.~\cite{DBLP:conf/edbt/Senin0WOGBCF15} converted time series subsequences into characters and used grammar rules to detect anomalies. In density-based methods, time series subsequences are grouped into clusters. Clusters with low density are then considered as anomalies. Breunig et al.~\cite{DBLP:conf/sigmod/BreunigKNS00} propose Local Outlier Factor (\texttt{LOF}), which considers the local density of clusters and is able to detect local outliers effectively. Sequeira and Zaki~\cite{DBLP:conf/kdd/SequeiraZ02} cluster time series subsequences into a fixed number of clusters using a $k$-medoids algorithm. Classical machine learning algorithms do not consider time series-specific temporal information, so they cannot be applied well in practical scenarios.

% \vspace{-5pt}
Deep learning based time series anomaly detection methods are used widely in many applications such as object monitoring~\cite{DBLP:conf/mm/ZhaoDSLLH17}, 
network analysis~\cite{DBLP:conf/icc/LuoN18},  robotics~\cite{DBLP:conf/icra/ParkEBK16}, and human behaviors analysis~\cite{DBLP:conf/mdm/Kieu0J18}. While diffusion-based models have recently shown impressive performance for generative models in terms of reconstruction quality, they are not intensively used in time series anomaly detection, and they also incur substantial computational costs. Crucially, our framework is orthogonal to the backbone choice. Thus, the proposed Encode-then-Decompose paradigm could be integrated with diffusion models. The latest methods include \texttt{AnomalyTrans}~\cite{DBLP:conf/iclr/XuWWL22}, which measures the strength of correlations between observations in time series, and \texttt{DCdetector}~\cite{DBLP:conf/kdd/YangZZW023}, which achieves impressive performance using a contrastive learning approach with a dual attention component. However, \texttt{AnomalyTrans} and \texttt{DCdetector} do not perform the encode-then-decompose mechanism like us. The most relevant study to our proposal is Robust Autoencoders (\texttt{RAE}s)~\cite{DBLP:conf/kdd/ZhouP17}, which decomposes a dataset into clean and anomalous components. The main difference between \texttt{RAE}s and \framework is that \texttt{RAE}s fail to handle temporal information and thus cannot work on time series. Further, \texttt{RAE}s decompose the data in the original space rather than in the latent representation space, as \framework does. \framework also integrates mutual information to better support decomposition. To the best of our knowledge, \framework is the first time series anomaly detection method that decomposes the latent variable to achieve robustness.

% \vspace{0.5em}
% \subsection{Mutual Information}
\noindent\textbf{Mutual Information.} Mutual information measures the relationship between statistical variables. Mutual information plays a role in many applications in a wide range of domains. Early approaches typically use nonparametric models for estimating mutual information~\cite{Kraskov_Stögbauer_Grassberger_2004}, such as kernel density estimation methods that use kernel functions to estimate the probability density function of data. Deep neural networks and representation learning~\cite{DBLP:journals/corr/abs-1807-03748} are being employed increasingly for mutual information estimation to cater to the demands posed by the expanding scale and complexity of contemporary datasets, as well as the need for representation optimization. Notable instances of this approach include  Barber-Agakov~\cite{DBLP:conf/nips/BarberA03}, mutual information neural estimator (\texttt{MINE})~\cite{DBLP:conf/icml/BelghaziBROBHC18}, and \texttt{M-estimators}~\cite{DBLP:journals/tit/NguyenWJ10}. Existing studies use mutual information to measure the relationship between variables in supervised learning problems where labeled data is available. To the best of our knowledge, our proposal is the first to use mutual information for unsupervised time series anomaly detection.

% \vspace{-0.5em}
\section{Conclusion}
\label{sec:con}
We propose \framework for unsupervised time series anomaly detection. The framework addresses a key problem in autoencoder-based anomaly detection methods: their high vulnerability to contaminated training data. The framework decomposes the latent representation into stable features and auxiliary features that comprise long-term patterns and point-wise patterns, respectively, rather than blindly reconstructing the time series. A mutual information criterion is integrated into the decomposition to support the robustness of the framework. Experimental studies show that the framework is effective and can outperform strong baselines and state-of-the-art methods. 

In future research, it is of interest to study anomaly detection in different settings, such as binary-value settings~\cite{DBLP:conf/kdd/YangZZW023}, semi-supervised settings~\cite{DBLP:conf/kdd/LiJDSL22}, time series of location-related information~\cite{DBLP:conf/icde/CirsteaYGKP22}, continual learning settings~\cite{DBLP:conf/icassp/WiewelY19}, and concept drift settings~\cite{DBLP:conf/cikm/TianKA0C19,DBLP:conf/nips/WangZQ0W0L23}. It is also of interest to study different approaches such as ensemble learning~\cite{DBLP:conf/ijcai/KieuYGJ19} and explainability~\cite{DBLP:conf/icde/KieuYGJZHZ22} to further improve anomaly detection accuracy.

\bibliographystyle{IEEEtranS}
\bibliography{ref}

\end{document}